\documentclass[runningheads]{llncs}
\usepackage{graphicx}
\usepackage{bbding}
\usepackage[dvipsnames]{xcolor}
\usepackage{tikz}
\usepackage{csquotes}
\usepackage{amsfonts}
\usepackage{bm}
\usepackage{wrapfig,lipsum,booktabs}
\usepackage{enumitem}
\usepackage{comment}
\usepackage{amsmath,amssymb} 
\usepackage{color}
\usepackage{booktabs}
\usepackage{graphicx}
\usepackage{caption}
\usepackage{subcaption}
\usepackage{multirow}

\begin{document}
\pagestyle{headings}
\mainmatter
\def\ECCVSubNumber{5076}  

\title{Factorizing Knowledge in  Neural Networks} 


%
\author{Xingyi Yang
\and Jingwen Ye
\and Xinchao Wang
}
%
\institute{National University of Singapore\\
\email{xyang@u.nus.edu,\{jingweny,xinchao\}@nus.edu.sg}}
\maketitle

\begin{abstract}
In this paper, we explore a novel and ambitious knowledge-transfer task, termed Knowledge Factorization~(KF). The core idea of KF lies in the modularization 
and assemblability of knowledge: given a pretrained network model as input, KF aims to decompose it into several factor networks, each of which handles only a dedicated task and maintains task-specific knowledge factorized from the source network. Such factor networks are task-wise disentangled and can be directly assembled, without any fine-tuning, to produce the more competent combined-task networks.  In other words, the factor networks serve as Lego-brick-like building blocks, allowing us to construct customized networks in a plug-and-play manner.  Specifically, each factor network comprises two modules, a common-knowledge module that is task-agnostic and shared by all factor networks, alongside with a task-specific module dedicated to the factor network itself.  
We introduce an information-theoretic objective, InfoMax-Bottleneck~(IMB), 
to carry out KF by 
optimizing the mutual information 
{between the learned representations and input}.
Experiments across various benchmarks demonstrate that, 
the derived factor networks yield gratifying performances 
on not only the dedicated tasks but also disentanglement, 
while enjoying much better interpretability and modularity. 
Moreover, the learned common-knowledge representations give rise to 
impressive results on transfer learning. Our code is available at \texttt{https://github.com/Adamdad/KnowledgeFactor}.

\end{abstract}

\section{Introduction}
\label{sec:intro}


Over the past decade, deep neural networks~(DNNs) 
have evolved to the \emph{de facto} 
a standard approach for most if not all 
computer vision tasks, yielding unprecedentedly promising
results. Due to the time- and resource-consuming
DNN training process, many developers have 
generously released their pretrained models online, 
so that users may adopt these models in a plug-and-play
manner without training from scratch. Nevertheless, 
pretrained DNNs often come with heavy architectures, 
making them extremely cumbersome to be deployed in 
real-world scenarios, especially resource-critical 
applications such as edge computing. 
Numerous endeavors have thus been made towards reducing the sizes of DNNs, among which one mainstream scheme is known as Knowledge Distillation~(KD). The goal of KD is to ``distill'' knowledge from a large pre-trained model known as a teacher, to a compact model known as a student. The derived student is expected to master the expertise of the teacher yet come with a much smaller size, making it applicable to edge devices. Since the seminal work of~\cite{Hinton2015DistillingTK}, a series of KD approaches have been proposed to strengthen the performances 
of student models~\cite{romero2014fitnets,Zagoruyko2017AT,pkt_eccv}.

\begin{figure*}[!t]
    \centering
    \includegraphics[width=1\linewidth]{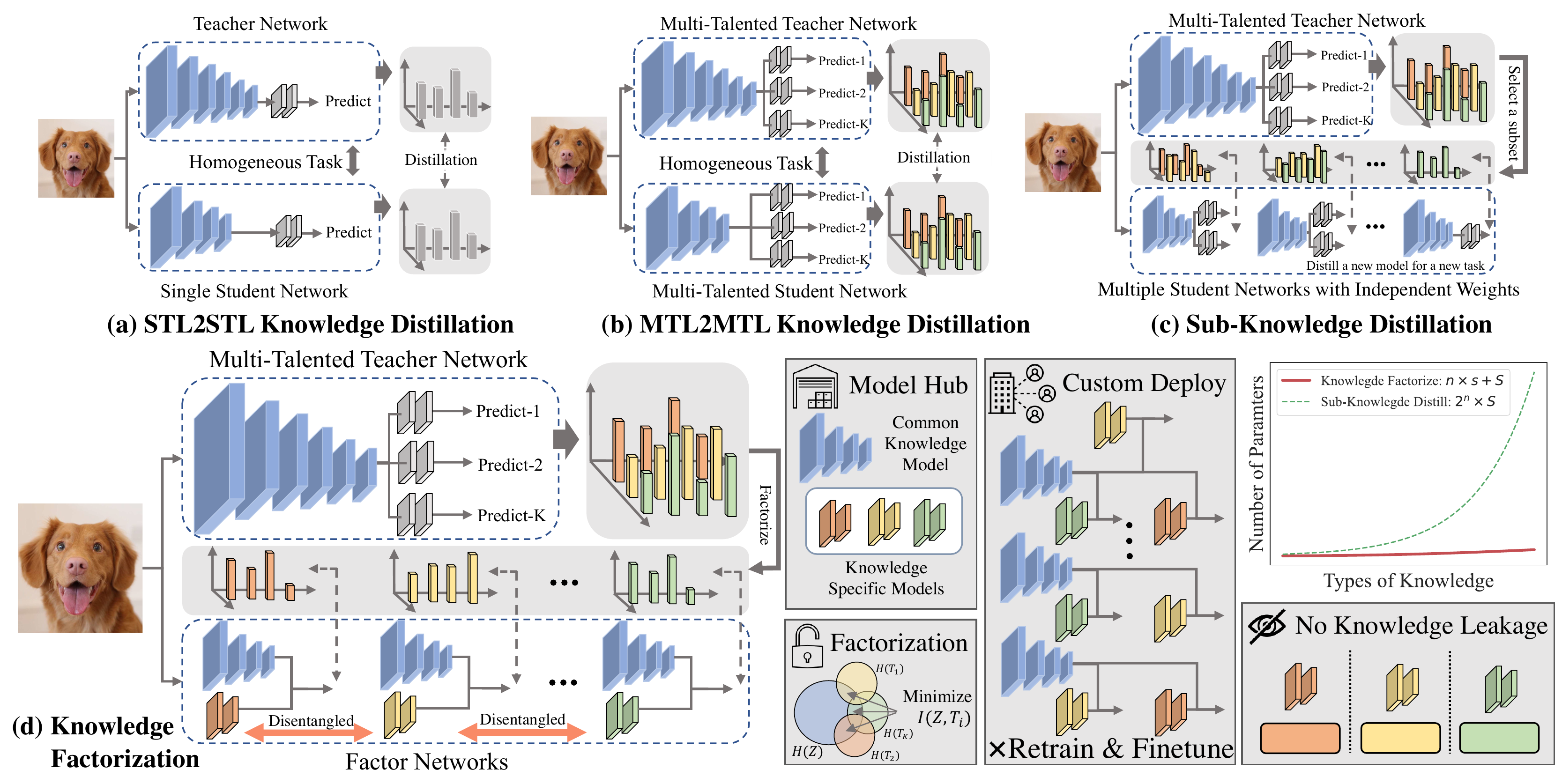}
    \caption{Illustration of (top) 3 types of Knowledge Distillation and (bottom) our proposed Knowledge Factorization. (a) Single-Task Learning to Single-Task Learning~(STL2STL) KD refers to distill a single-tasked student from a single-tasked
    teacher, (b) Multi-Task Learning to Multi-Task Learning~(MTL2MTL) KD stands for distilling a multi-tasked student from a multi-tasked
    teacher and (c) Sub-Knowldege Distillation distill a subset of the teacher's knowledge to its student model. 
    }
    \label{fig:KF_KD}
\end{figure*}

Albeit encouraging results achieved, KD has largely been treated as a black-box procedure, in which the intrinsic knowledge flow process remains opaque. Consequently, the derived student model may inherit the teacher's task-wise competence but unfortunately lacks interpretability, since it is unclear how and what knowledge has been transferred to the student. In addition, as demostrated in Fig~\ref{fig:KF_KD}(a) and (b), conventional KD assumes that teacher and student models master homogeneous tasks or knowledge, which greatly limits its wide applications. Even if it is allowed to distill a subset of knowledge from the teacher, shown in Fig~\ref{fig:KF_KD}(c), the problem setup of KD, by nature, overlooks the scalability of the student. For example, given a versatile classification teacher pretrained on ImageNet, if we are to learn two students, one handling cat-dog classification and one handling cat-fish, we will have to carry out the KD twice; if, however, we are to learn all $k$-class classification students from a pool of $1,000$ classes, 
we will have to conduct KD for $\sum_{k=1}^{1000} \binom{1,000}{k} = 2^{1000}$ times, which is computational intractable.

In this paper, we introduce a novel task, termed  Knowledge Factorization~(KF), that alleviates the aforementioned flaws of KD at a problem-setup level. The core idea of KF regards the modularization and assemblability of knowledge: given a pretrained teacher, KF decomposes it into several \emph{factor networks}, each of which masters one specific knowledge factorized from the teacher,
while remaining disentangled with respect to others. 
Moreover, these factor networks are expected to be readily integratable, meaning that
we may directly assemble multiple factor networks, without any fine-tuning, to produce a more competent multi-talented network. As shown in Fig.~\ref{fig:KF_KD}(d), those {factor networks} can be organized into a open-sourced model hub. At the same time, users could treat them as Lego-brick-like units of knowledge to build customized networks in a plug-and-play fashion, thereby lending itself to great scalability. 
Furthermore, the disentanglement property effectively
enables the IP protection of network knowledge:
since the factor networks are learned in a disentangled manner,
they possess only task-specific knowledge,
allowing the network owners to
selectively conduct knowledge transfer
without leaking knowledge of
other tasks.

Admittedly, the aims of KF are unarguably ambitious, 
since the factor networks are, again, expected to be modularized and readily integratable,
and meanwhile knowledge-wise disentangled and hence more interpretable.
Notably, despite orthogonal in expertise, 
these factor networks will inherit the common knowledge shared by all tasks. 
As such, each factor network should be designed to account for both the task-agnostic commonality and its task-relevant specialization, which in turn reduces the overall parameter overhead for KF. 
As demonstrated in Fig~\ref{fig:KF_KD}, given $n$ types of knowledge, sub-KD 
requires an exponential number of $2^n$ 
models, each with $S$ parameters, 
while KF reduces the model number to a linear scale, 
with one full-sized common knowledge model 
and $n$ mini models, each with $s$ parameter, where $s\ll S$. 

To this end, we propose a dedicated scheme for conducting KF,
that comprises two  mechanisms,
namely \emph{structural factorization} and \emph{representation factorization}.
\begin{itemize}
    \item{{\bf Structural Factorization.} 
    {Structural factorization decomposes the 
    teacher network into a set of factor
    networks with different functionalities. 
    Each factor network comprises a
    shared \textit{common-knowledge network}~(CKN)
    and a \textit{task-specific network}~(TSN). 
    CKN extracts task-agnostic representations 
    to capture the commonality among tasks, 
    whereas the TSN accounts for task-specific 
    information. Factor networks are trained to 
    specialize in an individual task via fusing 
    task-agnostic and task-specific knowledge.}
    }
    \item{{\bf Representation Factorization.}
    {Representation factorization disentangles 
    the shared knowledge and task-level 
    representations into statistically 
    independent components. For this purpose, 
    we introduce a novel information-theoretical objective,
    termed \textit{InfoMax Bottleneck}~(IMB). 
    It maximizes the mutual information between 
    input and the common features to encourage 
    the lossless information transmission in CKN. 
    Meanwhile, IMB minimizes data-task mutual 
    information to ensure that,  
    {the task features are only predictive for a specific task.}
    Specifically, we derive a variational lower 
    bound for IMB to practically optimize this loss.}
    }
\end{itemize}

By integrating both mechanisms,
we demonstrate in the experiments 
that KF indeed achieves architecture-level 
and representation-level 
disentanglement. Different from KD that transmits holistic knowledge in a black-box manner, 
KF offers unique interpretability 
for the factor networks through the knowledge transfer.
Moreover, the learned common-knowledge representations facilitate the transfer learning to unseen downstream tasks,
as will be verified empirically in our experiments.

Our contribution are therefore summarized as follows
\begin{itemize}
    \item We introduce a novel knowledge-transfer task, 
    termed \textit{Knowledge Factorization}~(KF), 
    which accounts for learning factor networks
    that are modularized and interpretable.
    Factor networks are expected to
    be readily integratable,
    without any retraining, to assemble multi-task
    networks.
    
    \item We propose an effective solution towards KF.
    Our approach decomposes a pretrained teacher into factor networks that are task-wise disentangled. 
    
    \item We design an \textit{InfoMax Bottleneck} objective to disentangle the representation between common knowledge and the task-specific representations, by exerting control over the mutual information between input and representations. We derive its variational bound
    for its numerical optimization.
    
    \item Our method achieves strong performance and disentanglement capability across various benchmarks, with better modularity and transferability.
\end{itemize}

\section{Related work}
\noindent\textbf{Knowledge Distillation.} Knowledge distillation~(KD)~\cite{Hinton2015DistillingTK} refers to the process to transfer the knowledge from one model or an ensemble of models to a student model. KD is originally designed for model compression~\cite{bucilua2006model,Sun2019PatientKD,li2020few,lopes2017data,yang2020CVPR,Sucheng2022CVPR,hu2019drnet}, but it has been found to be beneficial in other tasks like adversarial defense~\cite{papernot2016distillation}, domain adaptation~\cite{granger2020joint,nguyen2021unsupervised}, continual learning~\cite{li2017learning,zenke2017continual} and amalgamate the knowledge from multiple teachers~\cite{DBLP:conf/ijcai/LuoWFHTS19,ye2019student,jing2021amalgamating}. 
Different from the common KD methods that disseminates knowledge as a whole, we factorize the knowledge of a multi-talented teacher to factor networks with disentangled representations. \\\textbf{Disentangled representation learning.} It is often assumed that real-world observations should be controlled by factors. Therefore, a recent line of research argues the importance of finding disentangled variables in representation learning~\cite{bengio2013representation,peters2017elements,niemeyer2021giraffe,locatello2019challenging,yang2020NeurIPS,Feng2018NeurIPS} while providing invariance in learning~\cite{goodfellow2009measuring,achille2018emergence,jaiswal2018unsupervised}. The disentanglement are usually done through adversarial learning~\cite{tran2017disentangled,liu2018multi,NIPS2016_ef0917ea,Chen2016InfoGANIR} or variational auto-encoder \cite{higgins2016beta,burgess2018understanding,Kim2018DisentanglingBF}. In this work, we aim to disentangle the task-agnostic and task-related representation by optimizing the mutual information. \\
\textbf{InfoMax principle and Information bottleneck.} As one of the foundations of machine learning, information theory has promoted a series of learning algorithms. \textit{InfoMax}~\cite{linsker1988self} is a core principle of representation learning that encourages the mutual information should be maximized between multi-views or between representation and input. This principle gave birth to the recent trend on self-supervised learning~\cite{bachman2019learning,hjelm2018learning,tschannen2019mutual} and contrastive learning~\cite{oord2018representation,chen2020simple,he2020momentum,khosla2020supervised,tian2020makes,grill2020bootstrap}. On the contrary, \textit{Information Bottleneck}~(IB)~\cite{tishby2000information} aims to compress the representation while achieving realistic reconstruction results. 
In this study, we take a unified view of the two principles in multi-task learning. Infomax guarantees the learning of common knowledge across tasks, while IB promotes task-specific knowledge for an individual task.  \\\textbf{Multi-task learning.} Multi-task learning~(MTL) is designed to train models that handle multiple tasks by taking advantage of the common information among tasks. 
Some recent solutions explore on the decomposition between shared and task-specific processing~\cite{maninis2019attentive,kanakis2020reparameterizing,zhang2020side}. Unlike conventional methods, we decompose a pre-training model into knowledge modules according to tasks.

\vspace{-2mm}
\section{Method}

\begin{figure}[t]
    \centering
    \includegraphics[width=0.7\linewidth]{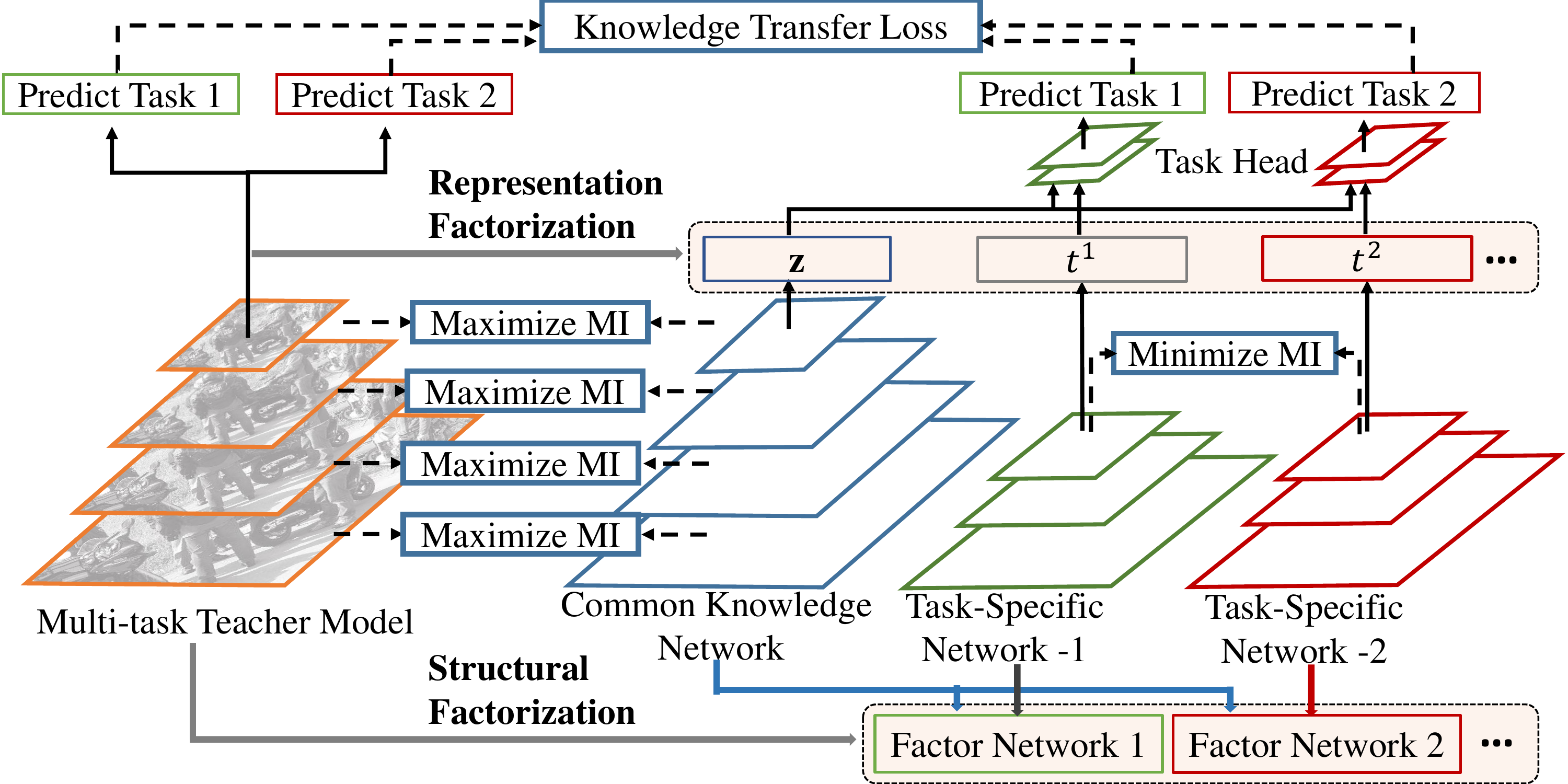}
    \caption{The overall framework of the proposed knowledge factorization. 
    The factor networks are trained to mimic the prediction of the teacher. The CKN learns to maximize the mutual information between input and its features, whereas the TSNs are dedicated to minimizing the task-wise mutual information.
    }
    \label{fig:KF with IMB}
\end{figure}

The essence of this work is to factorize a multi-task teacher into independent students by posing fine-grained control of the information among teacher and students. 
Figure~\ref{fig:KF with IMB} provides an overall sketch of our proposed KF. 
In what follows, we first give a definition of knowledge factorization, and then introduce the general procedure to decompose a teacher into factorized students. 
\subsection{Knowledge Factorization in Neural Network}
We define Knowledge Factorization~(KF) to
be the process of subdividing a teacher network 
into multiple \emph{factor networks},
each of which possesses distinctive knowledge 
to handle one task.
Formally, assume we have a multi-task 
dataset $\mathcal{D}=\{(\mathbf{x}_i,y_i^{1},\dots,y_i^{K})\}$, 
where each input sample $\mathbf{x}$ may take one of
$K$ different labels $\{y^j\}_{j=1}^K$ sampled from 
the joint probability $P(X, Y_1, \dots, Y_K)$. 
With a loose definition, we also deem the 
multi-classing as a special case for multi-tasking,
by considering each or a group of categories as a task. 
Given a multi-task teacher model
$\mathcal{T}$ that is able to predict $K$ 
tasks simultaneously, KF aims to construct $K$
factor networks $\{\mathcal{S}_j\}_{j=1}^K$,
each of which, again, tackles one task independently. 

Specifically, we focus on decomposing the teacher knowledge into task-specific and common representations, meaning that  each factor network not only masters task-specific knowledge, but also benefits from a shared common feature to make final predictions. 
{To this end,
we design two mechanisms to factorize knowledge: 
\emph{structural factorization}
to decompose the teacher network into
a set of factor networks 
, as well as \emph{representation factorization} 
to disentangle the common features from
task-specific features by optimizing  
mutual information.}

\subsection{Structural Factorization}
The goal of structural factorization is to 
endow different sub-networks with functional distinctions.
Each factor networks is expected to
inherit only a portion of 
the knowledge from the teacher,
and specializes in an individual task. 
Specifically, a factor network $\mathcal{S}_j$ for the $j$-th task comprises two modular networks: a Common Knowledge Network~(CKN) $\mathcal{S}_{C}(\cdot;\Theta_{\mathcal{S}_{C}})$ which is shared across all tasks, and a Task-specific Network~(TSN) $\mathcal{S}_{T_j}(\cdot;\Theta_{\mathcal{S}_{T_j}})$ which is task-exclusive. $\Theta_{\mathcal{S}_{C}}$ and $\Theta_{\mathcal{S}_{T_j}}$ are the model parameters for CKN and TSN respectively.
For each input sample, $\mathcal{S}_{C}$ is adopted to extract the task-agnostic feature $\mathbf{z}$: 
{\footnotesize
\setlength{\abovedisplayskip}{5pt}
\setlength{\belowdisplayskip}{5pt}
\begin{align}
    \mathbf{z} = \mathcal{S}_{C}(\mathbf{x};\Theta_{\mathcal{S}_{C}}).
\end{align}}\noindent
On the contrary, $\mathcal{S}_{T_j}$ learns the task-related knowledge $\mathbf{t}^j$ from the input $\mathbf{x}$, which together with $\mathbf{z}$ is processed by a task head $\mathcal{H}_{j}$ to make the final prediction:
{\small
\setlength{\abovedisplayskip}{5pt}
\setlength{\belowdisplayskip}{5pt}
\begin{align}
    \mathbf{t}^j = \mathcal{S}_{T_j}(\mathbf{x};\Theta_{\mathcal{S}_{T_j}});
    \hat{y}_S^j = \mathcal{H}_{j}(\mathbf{z}, \mathbf{t}^j;\Theta_{\mathcal{H}_{j}}),
\end{align}}\noindent
which constrains each factor network $\mathcal{S}_j$ to share the same common knowledge network but maintain the task-specific one to handle different tasks.

Intuitively, we expect that $\mathcal{S}_j$ only masters the knowledge about task $j$ by using the common knowledge $\mathbf{z}$ and $\mathbf{t}^j$. 
We accordingly define a structure factorization objective $\mathcal{L}_{sf}^{(j)}$ to enforce each single-task factor network to imitate the teacher's prediction while minimizing the supervised loss:
{\footnotesize\setlength{\abovedisplayskip}{5pt}
\setlength{\belowdisplayskip}{5pt}
\begin{align}
     \mathcal{L}_{sf}^{(j)} = \mathcal{L}^{(j)}_{\text{sup}} + \lambda_{\text{kt}}\mathcal{L}^{(j)}_{\text{kt}}, 
\end{align}}\noindent
where {$\mathcal{L}^{(j)}_{\text{sup}}$ and $\mathcal{L}^{(j)}_{\text{kt}}$} denote the supervised loss and the knowledge transfer loss for the $j$-th task, respectively,
and $\lambda_{\text{kt}}$ is the weight coefficient. 
Notably, we may readily adopt various implementations for
each of the loss terms here. For example,
$\mathcal{L}^{(j)}_{\text{sup}}$ may take the form
of L2 norm for regression and cross-entropy for classification,
while  $\mathcal{L}^{(j)}_{\text{kt}}$ may take the form
of of soft-target~\cite{Hinton2015DistillingTK}, hint-loss~\cite{romero2014fitnets}, 
or attention transfer~\cite{Zagoruyko2017AT}.
{More details can be found in the supplement.}


Structure factorization therefore
enables us to construct new combined-task models by assembling
multiple networks without retraining. 
If, for example, a 3-category classifier is needed, we can readily integrate CKN and the corresponding 3 TSNs from the pre-defined network pool. 
This property, in turn, greatly improves the scalability of the model.

\subsection{Representation Factorization}
Apart from the functionality disentanglement, 
we hope that learned representations 
of the factor networks are 
statistically independent as well,
so that each sub-network {masters task-wise disentangled knowledge}.
This means 
task-specific features should only contain minimal
information only related to a certain task, 
while the common representation contains as much information as possible. 

To this end, we 
introduce the \textit{Infomax Bottleneck}~(IMB) objective to 
optimize
the mutual information~(MI) between features and input.
For two random variables $X,Y$, MI $\mathcal{I}(X,Y)$ quantifies the ``number information'' that variable $X$ tells about $Y$, denoted by Kullback Leibler~(KL) divergence between the joint probability $p(\bm x, \bm y)$ and the product of marginal distribution $p(\bm x)p(\bm y)$:
 {\footnotesize\setlength{\abovedisplayskip}{5pt}
\setlength{\belowdisplayskip}{5pt}
\begin{equation}
    \begin{split}
            \mathcal{I}(X,Y) & = D_{KL}\Big[p(\bm x,\bm y)||p(\bm x)p(\bm y)\Big].
    \end{split}
\end{equation}}\noindent
In our problem, for each input sample $\mathbf{x}\sim P(X)$, we compute its common knowledge feature $\mathbf{z}\sim P(Z)$ and the task-predictive representation $\mathbf{t}^j \sim P(T_j)$. Ultimately, 
{IMB attempts to maximize $\mathcal{I}(X, Z)$
so that common knowledge keeps as much information of the input as possible,
while minimize $\mathcal{I}(X, T_j)$ so that task representations 
only preserve information related to the task.} 
{The representation disentanglement can then be formulated as an optimization problem:}
{\footnotesize\setlength{\abovedisplayskip}{5pt}
\setlength{\belowdisplayskip}{5pt}
\begin{equation}
\begin{split}
    \max  \mathcal{I}(T_j, Y_j); 
    \quad \text{s.t. } \mathcal{I}(X,T_j) \leq \epsilon_{1}, -\mathcal{I}(X, Z) \leq \epsilon_{2},
    \label{Eq: IMB}
\end{split}
\end{equation}}\noindent
where $\epsilon_1$ and $\epsilon_2$ are the information constraints we define. In order to solve Eq.~\ref{Eq: IMB}, we introduce two Lagrange multiplier $\alpha>0,\beta>0$ to construct the function:
{\footnotesize\setlength{\abovedisplayskip}{5pt}
\setlength{\belowdisplayskip}{5pt}
\begin{align}
    \mathcal{L}^{(j)}_I = \mathcal{I}(T_j, Y_j) + \alpha \mathcal{I}(X, Z) - \beta \mathcal{I}(X, T_j). 
\end{align}}\noindent
By maximizing the first term $\mathcal{I}(T_j, Y_j)$, we ensure that the task representation $\mathbf{t}^j$ is capable to accomplish individual task $j$. $\mathcal{I}(X, Z)$ term encourages the lossless transmission of information and high fidelity feature extraction for the CKN,
while minimizing $\mathcal{I}(X, T_j)$ enforces the only the task-informative representation is extracted by TSN, thus de-correlate the task knowledge $\mathbf{t}^j$ with the common knowledge $\mathbf{z}$. 
Unlike the convectional information bottleneck~(IB) principle~\cite{tishby2000information},
our proposed IMB attempts to 
maximize $\mathcal{I}(X, Z)$~\cite{hjelm2018learning,lowe2019greedy,oord2018representation}, so that the CKN learns a general representation $\mathbf{z}$ with high fidelity. 


\subsection{Variational Bound for Mutual Information}
Due to the difficulty of estimating mutual information for continuous variables, we derive a variational lower bound to approximate the exact IMB objective\footnote{
Due to space limitations, we only show the final formulations in the
main body of this paper. The derivations can be found in the supplementary
material.}:
{\footnotesize\setlength{\abovedisplayskip}{5pt}
\setlength{\belowdisplayskip}{5pt}
\begin{equation}
    \begin{split}
        \hat{\mathcal{L}_I} &= \mathbb{E}_{p(\mathbf{y}_j, \mathbf{t}_j)}[\log q(\mathbf{y}_j| \mathbf{t}_j)] + \alpha \big(\mathbb{E}_{p(\mathbf{z}, \mathbf{x})}[\log q(\mathbf{z}| \mathbf{x})] + H(Z)\big) - \beta \mathbb{E}_{p(\mathbf{x})}\Big[D_{KL}[p(\mathbf{t}_j|\mathbf{x})||q(\mathbf{t}_j)]\Big],
    \end{split} \label{eq: obj ifc}
\end{equation}}\noindent
where $D_{KL}$ denotes the KL divergence between two distributions and $q(\cdot)$ denotes the variational distributions. We claim that $\mathcal{L}_I \geq \hat{\mathcal{L}_I}$, with the equality achived if and only if $q(\mathbf{y}_j| \mathbf{t}_j) = p(\mathbf{y}_j| \mathbf{t}_j)$, $q(\mathbf{z}| \mathbf{x}) = p(\mathbf{z}| \mathbf{x})$ and $q(\mathbf{t}_j) = p(\mathbf{t}_j)$.

For better understanding, 
we explain the meaning of each term, specify the parametric forms of variational distribution and implementation details of Eq.~\ref{eq: obj ifc}.  \\
\textbf{Term 1.} We maximize $\mathcal{I}(T_j, Y_j)$ by maximizing its lower bound $\mathbb{E}_{p(\mathbf{y}_j, \mathbf{t}_j)}[\log q(\mathbf{y}_j| \mathbf{t}_j)]$. We set $q(\mathbf{y}_j|\mathbf{t}_j)$ to Gaussian for regression tasks and the multinomial distribution for classification tasks. Under this assumption, maximizing $\mathbb{E}_{p(\mathbf{y}_j, \mathbf{t}_j)}[\log q_(\mathbf{y}_j| \mathbf{t}_j)]$ is nothing more than minimizing the L2 norm or cross-entropy loss for the prediction. $q(\mathbf{y}_j|\mathbf{t}_j)$ is parameterized with another task head $\mathcal{H}_{j'}$ that takes $\mathbf{t}^j$ as input and makes the task prediction. Notably, $\mathcal{H}_{j'}$ is different from $\mathcal{H}_{j}$ since $\mathcal{H}_{j}$ takes both $\mathbf{z}$ and $\mathbf{t}^j$ as input.\\
\textbf{Term 2.} We maximize $\mathcal{I}(X, Z)$ by maximizing its lower bound $\mathbb{E}_{p(\mathbf{z}, \mathbf{x})}[\log q(\mathbf{z}| \mathbf{x})] + H(Z)$. We choose $q(\mathbf{z}| \mathbf{x})$ to be an energy-based function that is parameterized by a critic function $f(\mathbf{x}, \mathbf{z}) : \mathcal{X}\times \mathcal{Z} \to \mathbb{R}$
{\footnotesize\setlength{\abovedisplayskip}{5pt}
\setlength{\belowdisplayskip}{5pt}\begin{align}
        q(\mathbf{z}|\mathbf{x}) = \frac{p(\mathbf{z})}{C} e^{f(\mathbf{x},\mathbf{z})}, \text{where } C= \mathbb{E}_{p(\mathbf{z})}\big[e^{f(\mathbf{x},\mathbf{z})}\big].
\end{align}}\noindent
Substituting $q(\mathbf{z}|\mathbf{x})$ into the second term gives us an unnormalized lower bound:
{\footnotesize\setlength{\abovedisplayskip}{5pt}
\setlength{\belowdisplayskip}{5pt}\begin{align}
  \mathcal{I}(X, Z) 
    \geq\mathbb{E}_{p(\mathbf{z}, \mathbf{x})}[f(\mathbf{x},\mathbf{z})] -  \log\mathbb{E}_{p( \mathbf{x})}[C],
\end{align}}\noindent
The same bound is also mentioned in Mutual Information Neural Estimation (MINE)~\cite{belghazi2018mine}. Different from original MINE, in our implementation, we estimate the $\mathcal{I}(X, Z)$ through a feature-wise loss between teacher and students. With a slight abuse of notation, we refer $\mathbf{z}_\mathcal{T}=\mathcal{T}(\mathbf{x})_l \in \mathbb{R}^{d_\mathcal{T}}$ and $\mathbf{z}_\mathcal{C} = \mathcal{S}_\mathcal{C}(\mathbf{x})_l\in \mathbb{R}^{d_\mathcal{C}}$ as the intermediate feature vectors from teacher and CKN at the $l$-th layer. Given a pair of  $(\mathbf{z}_\mathcal{T}, \mathbf{z}_\mathcal{C})$, $f$ is defined as inner product of two vectors $f(\mathbf{x},\mathbf{z}_\mathcal{C}) = \langle {\mathbf{z}_\mathcal{C},FFN(\mathbf{z}_\mathcal{T})} \rangle $, where $FFN(\cdot):\mathbb{R}^{d_\mathcal{T}}\to \mathbb{R}^{d_\mathcal{C}}$ is a feed-forward network to align the dimensions between $\mathbf{z}_\mathcal{T}$ and $\mathbf{z}_\mathcal{C}$.  \\
\textbf{Term 3.} $\mathbb{E}_{p(\mathbf{x})}\big[D_{KL}[p(\mathbf{t}_j|\mathbf{x})||q(\mathbf{t}_j)]\big]$ is the expected KL divergence between the posterior $p(\mathbf{t}_j|\mathbf{x})$ and the prior $q(\mathbf{t}_j)$, which is a upper bound for $\mathcal{I}(X, T_j)$. We minimize $\mathcal{I}(X, T_j)$ by minimizing $\mathbb{E}_{p(\mathbf{x})}\big[D_{KL}[p(\mathbf{t}_j|\mathbf{x})||q(\mathbf{t}_j)]\big]$.

Following the common practice in variational inference~\cite{kingma2013auto,higgins2016beta}, we set the prior $q(\mathbf{t}_j)$ as zero-mean unit-variance gaussuian.  Besides, we assume the $p(\mathbf{t}_j|\mathbf{x})=\mathcal{N}(\bm \mu_{t_j}, \text{diag}(\bm \sigma_{t_j}) )$ is a Gaussian distribution. 
Accordingly, we compute the mean and variance for the task feature $\mathbf{t}_j$ in each forward pass:
{\footnotesize\setlength{\abovedisplayskip}{5pt}
\setlength{\belowdisplayskip}{5pt}
\begin{align}
    \mathbf{t}_j = \mathcal{S}_{T_j}(\mathbf{x}; \Theta_{\mathcal{S}_{T_j}});\bm\mu_{t_j}=\mathbb{E}[\mathbf{t}_j], \bm \sigma^2_{t_j}=\text{Var}[\mathbf{t}_j],
\end{align}}
\noindent
Then, the KL divergence between $p(\mathbf{t}_j|\mathbf{x})$ and $q(\mathbf{t}_j)$ can be computed as:
{\footnotesize
\setlength{\abovedisplayskip}{5pt}
\setlength{\belowdisplayskip}{5pt}
    \begin{align}
        D_{KL}[p(\mathbf{t}_j|\mathbf{x})||q(\mathbf{t}_j)] = \frac{1}{2}\sum_{l=1}^{L}(1 + \log \sigma_{t_j}^{(l)} -(\mu_{t_j}^{(l)})^2 - \sigma_{t_j}^{(l)}).
    \end{align}
    }\noindent
The superscript denotes the $l$-th element of $\bm \mu_{t_j}$ and $\bm \sigma_{t_j}$.
\subsubsection{Training.}
We minimize the following overall loss to achieve both structural and representation factorization between students:
{\footnotesize\setlength{\abovedisplayskip}{5pt}
\setlength{\belowdisplayskip}{5pt}
\begin{align}
    \min_{\Theta_{\mathcal{S}_C}, \Theta_{\mathcal{S}_{T_j}}, \Theta_{\mathcal{H}_j}} \sum_{j=1}^K \mathcal{L}_{sf}^{(j)} - \lambda_{\text{I}} \mathcal{L}^{(j)}_{\text{I}},
\end{align}}
 where $\lambda_{\text{I}}$ is weighting coefficent of the IMB objective. 

\section{Experiments}
In this section, we investigate how factorization works to promote the performance, modularity and transferability of the model. 
Defaultly, we set $\alpha$=1.0 and $\beta$=1e-3, $\lambda_{\text{I}}$=1 and $\lambda_{\text{kt}}$=0.1. 
Due to the space limit, 
more hyper-parameter settings, distillation loss,
implementation details, 
data descriptions, 
and definitions of the metrics 
are listed in supplementary material.


\subsection{Factor Networks Make Strong Task Prediction}
\begin{figure}[b]
    \centering
    \includegraphics[width=\linewidth]{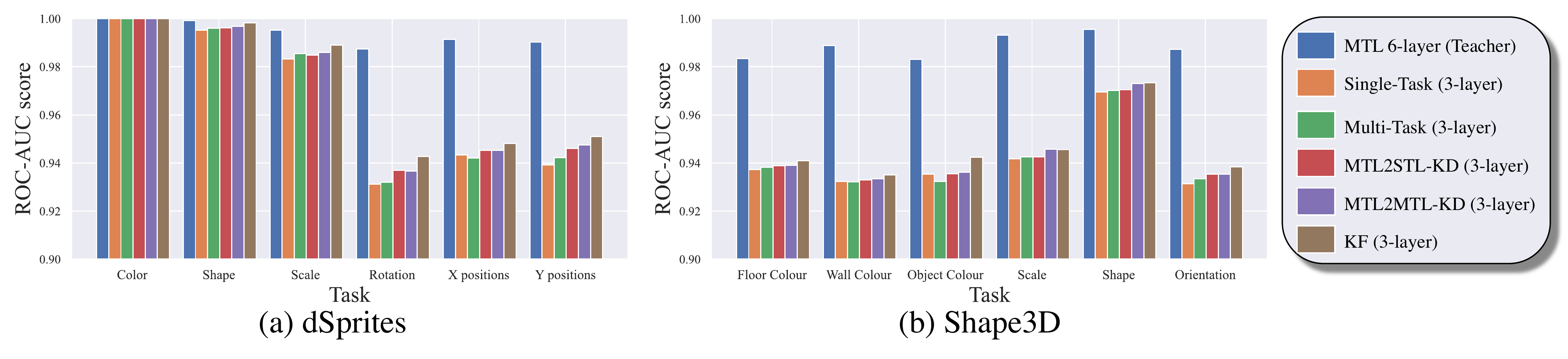}
        \vspace{-7mm}
    \caption{Test ROC-AUC comparison on dSprites and Shape3D datasets.}
    \label{fig:auc}
\end{figure}
We conduct comprehensive experiments 
on synthetic and real-world classification 
and multi-task benchmarks 
to investigate 
whether the factorized networks still maintain competitive predictive performance, especially on each subtask.
\noindent\textbf{Synthetic Evaluation.} We  first evaluate our KF
on two synthetic imagery benchmarks dSprites~\cite{dsprites17} and Shape3D~\cite{3dshapes18}. Two datasets 
are both generated 
by 6 ground truth independent latent factors. 
We define 
each latent factor 
as a prediction target 
and treat both datasets as multi-label classification benchmarks. We compare our KF with 4 other baseline methods: single-task baseline, multi-task baselines, MTL2MTL KD and MTL2STL KD.
Single-task baseline
denotes training 6 single-task networks,
while multi-task 
denotes that one model trained to predict
all 6 tasks. MTL2MTL KD distill a multi-tasked student, whereas
MTL2STL KD refers to 
distilling 6 single-tasked students. 
KF represents our results with factor networks. 
We train a teacher network as 6-layer CNN model. Besides, all students network encoders,
including both the CKN and TSNs, 
are parametrized by the 3-layer CNN.  We take a random train-test split of 7:3 on each dataset and report the ROC-AUC score on the test split.\\
\textbf{Results.} Figure~\ref{fig:auc} visualizes the bar plots for the ROC-AUC scores for our KF and its KD opponents on two datasets. Though all method achieves a high AUC score larger than $0.92$ on both datasets, it is evident that our KF not only surpasses the multi-tasked baseline but also exceeds two distillation paradigms. In addition, it is noted that multi-tasked models generally achieves better performance than their single-task counterpart, revealing that the prediction performance benefits from learning from multiple labels on two datasets.

\noindent\textbf{Real Image Classification.} We further evaluate our KF
on two real image classification CIFAR-10~\cite{krizhevsky2009learning} and ImageNet1K~\cite{ILSVRC15}.
To apply factorization, 
we construct two \textit{Pseudo-Multi-task Datasets} 
by considering the category hierarchy. 
The 10 classes in CIFAR-10 
can be divided into 6 \textit{animal} and 4 \textit{vehicle} categories. 
Similarly, ImageNet1K classes 
are organized using WordNet~\cite{miller1995wordnet} synset tree, with 11 super-classes.
We accordingly construct the CIFAR-10 2-task and ImageNet1K 11-task datasets, with each task considering one super-class. 

On the single-task 
and pseudo-multi-task evaluations,
we take a pretrained classifier 
and distill or factorize 
its knowledge to single-task 
or pseudo-multi-task students. 
Each pseudo-multi-task factor/distilled network 
only manages to predict the categories within one super-class, with the concatenated output serving as the final prediction. We include ResNet-18~\cite{he2016deep}, WideResNet28-2~(WRN28-2)~\cite{zagoruyko2016wide} and WideResNet28-10~(WRN28-10)~\cite{zagoruyko2016wide} as our teacher networks on CIFAR-10; MobileNetv2~(MBNv2)~\cite{sandler2018mobilenetv2}, along with ResNet-18, WRN28-2 as student or CKN backbone. On ImageNet1K evaluation, the teacher networks are selected to be ResNet-18, ResNet-34~\cite{he2016deep} and ResNet-50~\cite{he2016deep}, with MBNv2 and ResNet-18 as student or CKN backbone. We select a lightweight backbone MBNv2x0.5 to be TSNs. MBNv2x0.5 represents the width multiplier is 0.5.  
\renewcommand{\arraystretch}{0.7}
\begin{table}[!tb]
    \centering
    \scriptsize
    \caption{Test Accuracy (\%) comparison on CIFAR-10 between KD and KF. We report mean$\pm$std over 3 runs.}
    \label{tab:CIFAR-10}
    \begin{tabular}{c|c|cc|cc}
    \toprule
    Teacher:Acc & Student/CKN:Acc & 1-Task KD & 2-Task KD & 1-Task KF & 2-Task KF   \\
    \midrule
    \multirow{3}{*}{ResNet-18:94.54} & MBNv2:93.58 & 93.79\tiny{$\pm$0.17} & 92.59\tiny{$\pm$0.08} & 94.03\tiny{$\pm$0.23} & \textbf{94.41}\tiny{$\pm$0.05} \\
     & ResNet-18:94.54 & 94.72\tiny{$\pm$0.24} & 93.69\tiny{$\pm$0.11} & 95.04\tiny{$\pm$0.12} & \textbf{95.20}\tiny{$\pm$0.04} \\
     & WRN28-2:93.98 & 94.57\tiny{$\pm$0.13} & 93.71\tiny{$\pm$0.22} & \textbf{94.86}\tiny{$\pm$0.17} & 94.77\tiny{$\pm$0.06} \\\midrule
    \multirow{3}{*}{WRN28-2:93.98} & MBNv2:93.58 & 94.14\tiny{$\pm$0.08}
    & 94.10\tiny{$\pm$0.03}
    & 94.34\tiny{$\pm$0.14} & \textbf{94.56}\tiny{$\pm$0.10} \\
     & ResNet-18:94.54 & 94.75\tiny{$\pm$0.22} &
     94.22\tiny{$\pm$0.07} &95.03\tiny{$\pm$0.12} & \textbf{95.12}\tiny{$\pm$0.12} \\
     & WRN28-2:93.98 & 94.02\tiny{$\pm$0.07} &
     93.31\tiny{$\pm$0.12} & 94.59\tiny{$\pm$0.11} & \textbf{94.62}\tiny{$\pm$0.13} \\\midrule
    \multirow{3}{*}{WRN28-10:95.32} & MBNv2:93.58 & 
    94.47\tiny{$\pm$0.31} &
     94.10\tiny{$\pm$0.22} &
     94.80\tiny{$\pm$0.15} & \textbf{94.97}\tiny{$\pm$0.15} \\
     & ResNet-18:94.54 & 95.28\tiny{$\pm$0.14} &
     94.62\tiny{$\pm$0.09} &\textbf{95.40}\tiny{$\pm$0.08} & 95.32\tiny{$\pm$0.05} \\
     & WRN28-2:93.98 & 94.68\tiny{$\pm$0.14} &
     94.11\tiny{$\pm$0.26} & 94.80\tiny{$\pm$0.07} &
     \textbf{95.03}\tiny{$\pm$0.12} \\
    \bottomrule
    \end{tabular}
\end{table}
\begin{table}[t]
    \centering
    \scriptsize
    \caption{Top-1 Accuracy (\%) comparison on ImageNet.}
    \label{tab:ImageNet}
    \begin{tabular}{c|c|c|cc}
    \toprule
    Teacher:Acc & Student/CKN:Acc & 1-Task KD & 1-Task KF & 11-Task KF \\
    \midrule
    \multirow{2}{*}{ResNet-18:69.90} & MBNv2:71.86 & 72.15 & 72.20\textcolor{Green}{\tiny{(+0.05)}} & \textbf{72.52}\textcolor{Green}{\tiny{(+0.37)}} \\
     & ResNet-18:69.90 & 70.53 & 70.26\textcolor{Red}{\tiny{(-0.27)}} & \textbf{70.93}\textcolor{Green}{\tiny{(+0.40)}} \\\midrule
    \multirow{2}{*}{ResNet-34:73.62} & MBNv2:71.86 & 72.58 & 72.95\tiny{\textcolor{Green}{(+0.37)}} & \textbf{73.12}\tiny{\textcolor{Green}{(+0.54)}} \\
     & ResNet-18:69.90 & 70.82 & 70.98\tiny{\textcolor{Green}{(+0.16)}} & \textbf{72.13}\tiny{\textcolor{Green}{(+1.31)}} \\\midrule
    \multirow{2}{*}{ResNet-50:76.55} & MBNv2: 71.86 & 72.73 & 72.92\textcolor{Green}{\tiny{(+0.19)}} & \textbf{73.15}\textcolor{Green}{\tiny{(+0.42)}}\\
     & ResNet-18:69.90 & 71.12 & 71.14\textcolor{Green}{\tiny{(+0.02)}} & \textbf{72.20} \textcolor{Green}{\tiny{(+1.08)}} \\
    \bottomrule
    \end{tabular}
\end{table}
\\
\textbf{Results.} Table~\ref{tab:CIFAR-10} and Table~\ref{tab:ImageNet} provide the classification accuracy comparison between single-task or pseudo-multi-tasked KD and our proposed KF over 3 runs. Though both approaches improve the baselines under the single-task setting, we note that KD fails to improve the results on the pseudo-multi-tasked evaluation. We also do not report the 11-task KD results on ImageNet because the accuracy is generally lower than 20\%. Notably, we observed that the imbalanced labeling causes the deterioration in training: when one network only masters one super-class and the rest of the classes are treated as negative samples, the distilled networks are prone to make low-confident predictions in the end. In comparison, KF has a CKN shared across all 
tasks, which considerably alleviates the imbalance problem in conventional KD. For example, factor networks obtained by 11-Task KF improve the performance of ResNet18-KD on ImageNet over 1.08\% and 1.31\% when learning from ResNet-50 and ResNet-34. On other evaluations, KF consistently makes progress overall the normal KD, which suggests that the factorization of task-specific and task-agnostic benefit the performance.

\noindent\textbf{Multi-Task Dense Prediction.} Two multi-task dense prediction datasets are also used to verify the effectiveness of KF, including NYU Depth Dataset V2~(NYUDv2)~\cite{silberman2012indoor} and PASCAL Context~\cite{chen2014detect}.
NYUDv2 dataset contains indoor scene images annotated for segmentation and monocular depth estimation. 
We include 4 tasks in PASCAL Context, 
including semantic/human part segmentation, normal prediction, 
and saliency detection. 
We use the mean intersection over union (mIoU), the angle mean error (mErr) and root mean square error (rmse) are 
used to measure the prediction quality.

We include both the single-task and multi-task together with their STL2STL/\\MTL2STL/MTL2MTL distilled models as our baselines. We adopt the  HRNet48~\cite{wang2020deep} and ResNet-50 DeepLabv3 as teacher and HRNet18 and ResNet-18 DeepLabv3 as student or CNK. The TSN are set to MBNv2x0.5. We use a smaller $\beta$=1e-5. The networks are initialized with the ImageNet pretrained weights.
\renewcommand{\arraystretch}{0.7}
\begin{table}[b]
    \centering
    \scriptsize
     \caption{Performance comparison on the NYUDv2 dataset.}
    \begin{tabular}{l|cc|c|c}
    \toprule
    Method & Teacher & Student/CKN & Seg.(mIoU)$\uparrow$  & Depth(rmse)$\downarrow$ \\
    \midrule
        Single-task & - & HRNet18 & 27.37 & 0.612 \\
        Multi-task &  - & HRNet18 & 37.59 & 0.641 \\
        Single-task & - & HRNet48 & 48.19 & 0.556\\
        Multi-task &  - & HRNet48 & 48.92 & 0.578 \\\midrule
        STL2STL-KD & HRNet48 & HRNet18 & 39.27 &  0.603 \\
        MTL2MTL-KD &  HRNet48 & HRNet18 & 38.02 &  0.604 \\
        MTL2STL-KD &  HRNet48 & HRNet18 & 39.04 &  0.601 \\
        Ours & HRNet48 & HRNet18 & \textbf{40.78} & \textbf{0.592} \\\midrule\midrule
        Single-task & - & ResNet-18 & 38.07 & 0.652 \\
        Multi-task &  - & ResNet-18 & 39.18 & 0.623 \\
        Single-task & - & ResNet-50 & 44.30 & 0.625 \\
        Multi-task &  - & ResNet-50 & 44.78 & 0.602 \\\midrule
        STL2STL-KD &  ResNet-50 & ResNet-18 & 39.76 &  0.633 \\
        MTL2MTL-KD &  ResNet-50 & ResNet-18 & 39.98 &  0.623 \\
        MTL2STL-KD &  ResNet-50 & ResNet-18 & 40.60 &  0.621 \\
        Ours & ResNet-50 & ResNet-18 & \textbf{41.33} & \textbf{0.615} \\
        \bottomrule
    \end{tabular}
    \label{tab:NYUDv2}
\end{table}
\begin{table}
    \centering
     \scriptsize
    \caption{Performance comparison on the PASCAL dataset.}
    \resizebox{\linewidth}{!}{
    \begin{tabular}{l|cc|c|c|c|c}
    \toprule
    Method & Teacher & Student/CKN  & Seg.(mIoU)$\uparrow$ &  H.Part(mIOU)$\uparrow$ & Norm.(mErr)$\downarrow$ & Sal.(mIOU)$\uparrow$  \\
    \midrule
        Single-task & - & HRNet18 & 51.18 & 64.10 & 14.54 & 56.08\\
        Multi-task &  - & HRNet18 & 54.61 & 62.40 & 14.77 & 66.07  \\
        Single-task & - & HRNet48 & 60.92 & 67.15 & 14.53 & 68.12 \\
        Multi-task &  - & HRNet48 & 55.93 & 67.06 & 14.31 & 67.08 \\\midrule
        STL2STL-KD &  HRNet48 & HRNet18 & 52.63 & 64.98 & 14.49 & 60.72  \\
        MTL2MTL-KD &  HRNet48 & HRNet18 & 52.02 & 60.33 & 14.63 & 65.45 \\
        MTL2STL-KD &  HRNet48 & HRNet18 & 54.77 & 65.18 & 14.53 & 64.31 \\
        Ours & HRNet48 & HRNet18 & \textbf{56.65} & \textbf{66.83} & \textbf{14.44} & \textbf{67.05} \\\midrule\midrule
        Single-task & - & ResNet-18 & 64.75 & 58.68 & 13.95 & 65.59 \\
        Multi-task &  - & ResNet-18 & 63.48 & 58.17  & 15.12 & 64.50\\
        Single-task & - & ResNet-50 & 70.29 & 61.47 & 14.65 & 66.22 \\
        Multi-task &  - & ResNet-50 & 68.04 &63.05 & 14.88 & 65.65\\\midrule
        STL2STL-KD &  ResNet-50 & ResNet-18 & 66.10 & 59.43 & \textbf{14.19} & 66.33 \\
        MTL2MTL-KD &  ResNet-50 & ResNet-18 & 61.31 & 60.14 & 14.73 & 62.45  \\
        MTL2STL-KD &  ResNet-50 & ResNet-18 & 66.60 & \textbf{62.33} & 14.29 & 66.14  \\
        Ours & ResNet-50 & ResNet-18 & \textbf{67.18} & 61.09 & 14.31 & \textbf{66.83} \\
        \bottomrule
    \end{tabular}}
    \label{tab:PASCAL}
\end{table}\\
\textbf{Results.} We show the evaluation results on NYUDv2 and PASCAL datasets in Table~\ref{tab:NYUDv2} and Table~\ref{tab:PASCAL}. On NYUDv2, the multi-task baselines are generally better-performed than its single-task competitors. On the contrary, in the PASCAL experiments of HRNet48, ResNet18 and ResNet50, the performance of multitask baseline has largely degraded. It reveals the \textit{negative transfer} problem in MTL that the joint optimization of multiple objective might cause the contradiction between tasks, thus leading to undesirable performance reduction. 

The same problem remains when comparing MTL2MTL-KD to STL2\\-STL-KD in Table~\ref{tab:PASCAL}, where the MTL teacher is inferior to STL ones. Our factor networks automatically resolve this problem, because different TSNs are structurally and representationally independent. As a result, KF achieved strong student performance compared to other baselines.


\subsection{Factorization brings Disentanglement}

Given the distilled and factorized models in the previous section we measure a set of disentanglement metrics and representation similarity
to confirm 
that the knowledge factorization 
captures the independent 
variables across tasks.\\
\textbf{Disentanglement Evaluation Setup.} 
We first validate the disentanglement between factor models
on dSprites~\cite{dsprites17} and Shape3D~\cite{3dshapes18}. 
We measure 4 disentanglement metrics 
to quantify how well the learned representations summarize the factor variables. 
Those metrics are disenanglement-completness-informativeness (DCI)~\cite{eastwood2018framework}, Mutual information gap (MIG)~\cite{chen2018isolating}, FactorVAE metric~\cite{Kim2018DisentanglingBF}, and Separated Attribute Predictability (SAP) score~\cite{kumar2017variational}. Higher means better. 

\begin{figure}[t]
    \centering
    \includegraphics[width=\linewidth]{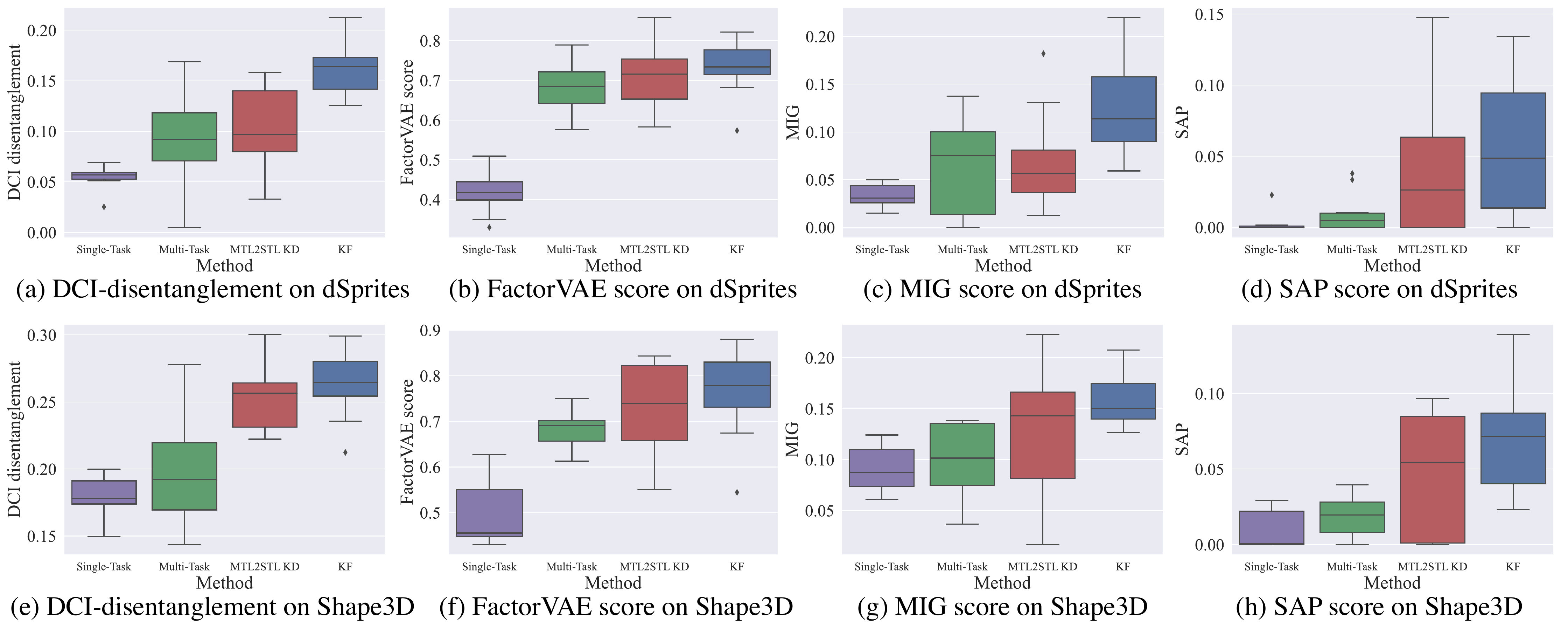}
    \caption{Disentanglement Metrics comparison between (1) Single-Task Baseline, (2) Multi-Task Baseline, (3) KD, and (4) our proposed KF on dSprite~(top) and Shape3D~(bottom) datasets. Each experiment is repeated over 10 runs. }
    \label{fig:disentangle metric}
\end{figure}

We compare our KF with 3 other baseline methods: single-task baseline, multi-task baselines, and MTL2STL KD students, which has been introduced in previous section.
Following the evaluation protocol in~\cite{locatello2019challenging}, we adopt 
the concatenation of all averge-pooled task-specific representations 
as our final feature vector for evaluation and compute all scores on test set.  \\
\textbf{Results.} Figure~\ref{fig:disentangle metric} illustrates the quantitative results
of different disentanglement metrics 
using box plots. 
First, we see that 
multi-task learning naturally comes with disentangled representations, where MTL achieves 
a slightly higher score 
than the STL.  Another observation is that 
knowledge transfer methods 
like KD and KF 
also help the model 
to find factors that are unappreciable for the teachers. 
The features extracted by our factor networks generally score the best, especially on the dSprites dataset, with an improvement over median of 0.47 and 0.09 on DCI and MIG scores. 
It is in line with our expectation that 
decomposing the knowledge into parts 
leads to disentangled representations.

\noindent\textbf{Representation Similarity.} We further conduct representation similarity analysis using centered kernel alignment~(CKA)~\cite{kornblith2019similarity} between teacher models, distilled models and our factorized models across 4 datasets, including dSprites, Shape3D, CIFAR10 and NYUDv2. On each dataset, CKA is adopted to quantifying feature similarity among (1) MTL teacher (2) MTL2MTL-KD student (3) MTL2STL students and (4) Our CKN and TSNs. We compute linear kernel CKA  between all pairs of models at the last feature layer on test set. The model architectures are described in the Appendix. The higher CKA index suggests higher correlation between two networks.
\begin{figure}[t]
    \centering
    \includegraphics[width=\linewidth]{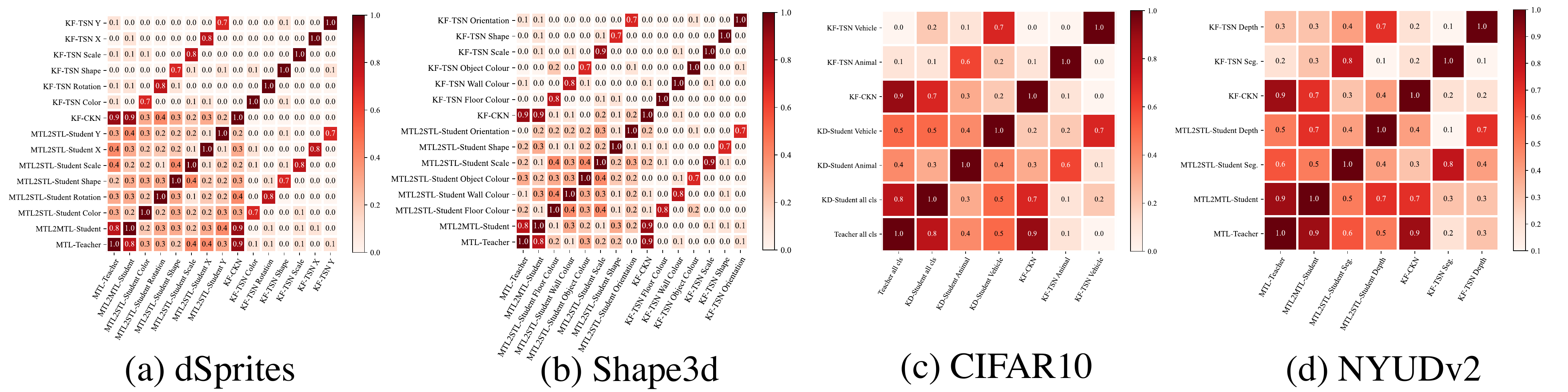}
    \caption{CKA Representation similarity between distilled and factorized models.}
    \label{fig:cka}
\end{figure}\\
\textbf{Results.} Figure~\ref{fig:cka} visualizes the CKA confusion matrix between all model pairs on 4 tasks. We made the following observations. First, models mastering the same subtask has high feature similarity.  Second, our factorized TSN captures more ``pure'' knowledge compared with MTL2STL students. On each heatmap, the bottom left region has high similarity~(in darker red), suggesting that the conventional distilled models still maintains high similarity with its peers even though they are trained on dedicated tasks. In comparison, factorized TSNs achieve smaller  similarities (in upper right region), again supporting our argument that factor networks capture the disentangled factors across tasks.
\subsection{Common Knowledge benefits Transferring}
We then finetune the factorized CKN on two downstream tasks 
to see if the common knowledge
facilities the transfer learning 
to unseen domains. 
We train ResNet-18 networks 
with different initializations 
on Caltech-UCSD Birds~(CUB-200)~\cite{WahCUB_200_2011} 
and MIT indoor scene~(Scene)~\cite{quattoni2009recognizing}. 
The trained models are then 
reestablished as teachers 
to educate student networks like MBNv2 and ShuffleNetv2. 
\\
\textbf{Results.} Table~\ref{tab:common transfer} shows 
the transfer learning performance 
and distillation accuracy 
using different pretrained weights. 
R18 w/ ImageNet-CKN refers to the ResNet-18 CKN
factorized from ImageNet pretrained ResNet-18. Compared with the original pretrained weights,
ImageNet-CKN achieves 
substantial improvement on both datasets.
By reusing the finetuned ResNet-18 as teacher network, we show in Figure~\ref{tab:common transfer} that CKN serves as a better role model 
to educate the student networks. 
It provides compelling evidence that common knowledge factorized from the teacher network benefits the transfer learning to other tasks. 

\begin{table}
    \centering
    \scriptsize
    \caption{Finetuning performance and distillation accuracy with different pretrained weights. R18 is the short for ResNet-18.}
    \begin{tabular}{l|l|c|c}
    \toprule
        Teacher & Student& CUB-200 & Scene\\
        \midrule
        - & R18 w/ Rand init. & 46.14 & 65.17 \\
        - & R18  w/ ImageNet & 65.28& 65.19 \\
        - & R18 w/ ImageNet-CKN & \textbf{69.17}& \textbf{72.37}\\
        \midrule\midrule
      - & MobileNetV2 w/ Rand init.& 48.80 & 64.59 \\
        R18 w/ Rand init. & MobileNetV2 w/ Rand init.& 54.18 &  66.78 \\
        R18 w/ ImageNet & MobileNetV2 w/ Rand init.& 61.30 & 66.40\\
        R18 w/ ImageNet-CKN & MobileNetV2 w/ Rand init.& \textbf{64.25} & \textbf{70.94}\\
         \midrule
      - & ShuffleNetv2 w/ Rand init.& 52.51 & 64.39 \\
        R18 w/ Rand init. & ShuffleNetv2 w/ Rand init.& 48.19 & 65.70 \\
        R18 w/ ImageNet & ShuffleNetv2 w/ Rand init.& 59.15 & 66.00 \\
        R18 w/ ImageNet-CKN & ShuffleNetv2 w/ Rand init.& \textbf{60.69} & \textbf{68.95}\\
    \bottomrule
    \end{tabular}
    \label{tab:common transfer}
\end{table}

\section{Conclusion} \vspace{-1mm}
In this paper, 
we introduce a novel knowledge-transfer task 
termed \textit{Knowledge Factorization}. 
Given a pretrained teacher, 
KF decomposes it into task-disentangled factor networks, 
each of which masters the task-specific 
and the common knowledge 
factorized from the teacher. 
Factor networks may operate independently, 
or be integrated to assemble multi-task networks, allowing for great scalability.
We design an InfoMax Bottleneck objective 
to disentangle the common and task-specific representations 
by optimizing the mutual information between input and representations. 
Our method 
achieves strong and robust performance, 
and meanwhile demonstrates great disentanglement capability across various benchmarks, with better modularity and transferability.
\vspace{-3mm}
\section*{Acknowledgement}
This work is supported by NUS Advanced Research and Technology Innovation Centre (ARTIC) Project Reference ECT-RP2, and Faculty Research Committee Grant (WBS: A-0009440-00-00).
Xinchao Wang is the corresponding author.

\clearpage
%
%
\bibliographystyle{splncs04}
\bibliography{egbib}

\begin{thebibliography}{10}
\providecommand{\url}[1]{\texttt{#1}}
\providecommand{\urlprefix}{URL }
\providecommand{\doi}[1]{https://doi.org/#1}

\bibitem{achille2018emergence}
Achille, A., Soatto, S.: Emergence of invariance and disentanglement in deep
  representations. The Journal of Machine Learning Research  \textbf{19}(1),
  1947--1980 (2018)

\bibitem{bachman2019learning}
Bachman, P., Hjelm, R.D., Buchwalter, W.: Learning representations by
  maximizing mutual information across views. arXiv preprint arXiv:1906.00910
  (2019)

\bibitem{belghazi2018mine}
Belghazi, M.I., Baratin, A., Rajeswar, S., Ozair, S., Bengio, Y., Courville,
  A., Hjelm, R.D.: Mine: mutual information neural estimation. arXiv preprint
  arXiv:1801.04062  (2018)

\bibitem{bengio2013representation}
Bengio, Y., Courville, A., Vincent, P.: Representation learning: A review and
  new perspectives. IEEE transactions on pattern analysis and machine
  intelligence  \textbf{35}(8),  1798--1828 (2013)

\bibitem{bucilua2006model}
Buciluǎ, C., Caruana, R., Niculescu-Mizil, A.: Model compression. In:
  Proceedings of the 12th ACM SIGKDD international conference on Knowledge
  discovery and data mining. pp. 535--541 (2006)

\bibitem{3dshapes18}
Burgess, C., Kim, H.: 3d shapes dataset.
  https://github.com/deepmind/3dshapes-dataset/ (2018)

\bibitem{burgess2018understanding}
Burgess, C.P., Higgins, I., Pal, A., Matthey, L., Watters, N., Desjardins, G.,
  Lerchner, A.: Understanding disentangling in $ \beta$-vae. arXiv preprint
  arXiv:1804.03599  (2018)

\bibitem{chen2018isolating}
Chen, R.T., Li, X., Grosse, R., Duvenaud, D.: Isolating sources of
  disentanglement in variational autoencoders. arXiv preprint arXiv:1802.04942
  (2018)

\bibitem{chen2020simple}
Chen, T., Kornblith, S., Norouzi, M., Hinton, G.: A simple framework for
  contrastive learning of visual representations. In: International conference
  on machine learning. pp. 1597--1607. PMLR (2020)

\bibitem{Chen2016InfoGANIR}
Chen, X., Duan, Y., Houthooft, R., Schulman, J., Sutskever, I., Abbeel, P.:
  Infogan: Interpretable representation learning by information maximizing
  generative adversarial nets. In: NIPS (2016)

\bibitem{chen2014detect}
Chen, X., Mottaghi, R., Liu, X., Fidler, S., Urtasun, R., Yuille, A.: Detect
  what you can: Detecting and representing objects using holistic models and
  body parts. In: Proceedings of the IEEE conference on computer vision and
  pattern recognition. pp. 1971--1978 (2014)

\bibitem{eastwood2018framework}
Eastwood, C., Williams, C.K.: A framework for the quantitative evaluation of
  disentangled representations. In: International Conference on Learning
  Representations (2018)

\bibitem{Feng2018NeurIPS}
Feng, Z., Wang, X., Ke, C., Zeng, A., Tao, D., Song, M.: Dual swap
  disentangling. In: Conference on Neural Information Processing Systems (2018)

\bibitem{goodfellow2009measuring}
Goodfellow, I., Lee, H., Le, Q., Saxe, A., Ng, A.: Measuring invariances in
  deep networks. Advances in neural information processing systems
  \textbf{22},  646--654 (2009)

\bibitem{granger2020joint}
Granger, E., Kiran, M., Dolz, J., Blais-Morin, L.A., et~al.: Joint progressive
  knowledge distillation and unsupervised domain adaptation. In: 2020
  International Joint Conference on Neural Networks (IJCNN). pp.~1--8. IEEE
  (2020)

\bibitem{grill2020bootstrap}
Grill, J.B., Strub, F., Altch{\'e}, F., Tallec, C., Richemond, P.H.,
  Buchatskaya, E., Doersch, C., Pires, B.A., Guo, Z.D., Azar, M.G., et~al.:
  Bootstrap your own latent: A new approach to self-supervised learning. arXiv
  preprint arXiv:2006.07733  (2020)

\bibitem{he2020momentum}
He, K., Fan, H., Wu, Y., Xie, S., Girshick, R.: Momentum contrast for
  unsupervised visual representation learning. In: Proceedings of the IEEE/CVF
  Conference on Computer Vision and Pattern Recognition. pp. 9729--9738 (2020)

\bibitem{he2016deep}
He, K., Zhang, X., Ren, S., Sun, J.: Deep residual learning for image
  recognition. In: Proceedings of the IEEE conference on computer vision and
  pattern recognition. pp. 770--778 (2016)

\bibitem{higgins2016beta}
Higgins, I., Matthey, L., Pal, A., Burgess, C., Glorot, X., Botvinick, M.,
  Mohamed, S., Lerchner, A.: beta-vae: Learning basic visual concepts with a
  constrained variational framework  (2016)

\bibitem{Hinton2015DistillingTK}
Hinton, G.E., Vinyals, O., Dean, J.: Distilling the knowledge in a neural
  network. ArXiv  \textbf{abs/1503.02531} (2015)

\bibitem{hjelm2018learning}
Hjelm, R.D., Fedorov, A., Lavoie-Marchildon, S., Grewal, K., Bachman, P.,
  Trischler, A., Bengio, Y.: Learning deep representations by mutual
  information estimation and maximization. arXiv preprint arXiv:1808.06670
  (2018)

\bibitem{hu2019drnet}
Hu, X., An, Z., Yang, C., Zhu, H., Xu, K., Xu, Y.: Drnet: Dissect and
  reconstruct the convolutional neural network via interpretable manners. arXiv
  preprint arXiv:1911.08691  (2019)

\bibitem{jaiswal2018unsupervised}
Jaiswal, A., Wu, Y., AbdAlmageed, W., Natarajan, P.: Unsupervised adversarial
  invariance. arXiv preprint arXiv:1809.10083  (2018)

\bibitem{jing2021amalgamating}
Jing, Y., Yang, Y., Wang, X., Song, M., Tao, D.: Amalgamating knowledge from
  heterogeneous graph neural networks. In: Proceedings of the IEEE/CVF
  Conference on Computer Vision and Pattern Recognition. pp. 15709--15718
  (2021)

\bibitem{kanakis2020reparameterizing}
Kanakis, M., Bruggemann, D., Saha, S., Georgoulis, S., Obukhov, A., Gool, L.V.:
  Reparameterizing convolutions for incremental multi-task learning without
  task interference. In: European Conference on Computer Vision. pp. 689--707.
  Springer (2020)

\bibitem{khosla2020supervised}
Khosla, P., Teterwak, P., Wang, C., Sarna, A., Tian, Y., Isola, P., Maschinot,
  A., Liu, C., Krishnan, D.: Supervised contrastive learning. arXiv preprint
  arXiv:2004.11362  (2020)

\bibitem{Kim2018DisentanglingBF}
Kim, H., Mnih, A.: Disentangling by factorising. ArXiv  \textbf{abs/1802.05983}
  (2018)

\bibitem{kingma2013auto}
Kingma, D.P., Welling, M.: Auto-encoding variational bayes. arXiv preprint
  arXiv:1312.6114  (2013)

\bibitem{kornblith2019similarity}
Kornblith, S., Norouzi, M., Lee, H., Hinton, G.: Similarity of neural network
  representations revisited. In: International Conference on Machine Learning.
  pp. 3519--3529. PMLR (2019)

\bibitem{krizhevsky2009learning}
Krizhevsky, A., Hinton, G., et~al.: Learning multiple layers of features from
  tiny images  (2009)

\bibitem{kumar2017variational}
Kumar, A., Sattigeri, P., Balakrishnan, A.: Variational inference of
  disentangled latent concepts from unlabeled observations. arXiv preprint
  arXiv:1711.00848  (2017)

\bibitem{li2020few}
Li, T., Li, J., Liu, Z., Zhang, C.: Few sample knowledge distillation for
  efficient network compression. In: Proceedings of the IEEE/CVF Conference on
  Computer Vision and Pattern Recognition. pp. 14639--14647 (2020)

\bibitem{li2017learning}
Li, Z., Hoiem, D.: Learning without forgetting. IEEE transactions on pattern
  analysis and machine intelligence  \textbf{40}(12),  2935--2947 (2017)

\bibitem{linsker1988self}
Linsker, R.: Self-organization in a perceptual network. Computer
  \textbf{21}(3),  105--117 (1988)

\bibitem{liu2018multi}
Liu, Y., Wang, Z., Jin, H., Wassell, I.: Multi-task adversarial network for
  disentangled feature learning. In: Proceedings of the IEEE Conference on
  Computer Vision and Pattern Recognition. pp. 3743--3751 (2018)

\bibitem{locatello2019challenging}
Locatello, F., Bauer, S., Lucic, M., Raetsch, G., Gelly, S., Sch{\"o}lkopf, B.,
  Bachem, O.: Challenging common assumptions in the unsupervised learning of
  disentangled representations. In: international conference on machine
  learning. pp. 4114--4124. PMLR (2019)

\bibitem{lopes2017data}
Lopes, R.G., Fenu, S., Starner, T.: Data-free knowledge distillation for deep
  neural networks. arXiv preprint arXiv:1710.07535  (2017)

\bibitem{lowe2019greedy}
L{\"o}we, S., O’Connor, P., Veeling, B.S.: Greedy infomax for self-supervised
  representation learning  (2019)

\bibitem{DBLP:conf/ijcai/LuoWFHTS19}
Luo, S., Wang, X., Fang, G., Hu, Y., Tao, D., Song, M.: Knowledge amalgamation
  from heterogeneous networks by common feature learning. In: Kraus, S. (ed.)
  Proceedings of the Twenty-Eighth International Joint Conference on Artificial
  Intelligence, {IJCAI} 2019, Macao, China, August 10-16, 2019. pp. 3087--3093.
  ijcai.org (2019). \doi{10.24963/ijcai.2019/428},
  \url{https://doi.org/10.24963/ijcai.2019/428}

\bibitem{maninis2019attentive}
Maninis, K.K., Radosavovic, I., Kokkinos, I.: Attentive single-tasking of
  multiple tasks. In: Proceedings of the IEEE/CVF Conference on Computer Vision
  and Pattern Recognition. pp. 1851--1860 (2019)

\bibitem{NIPS2016_ef0917ea}
Mathieu, M.F., Zhao, J.J., Zhao, J., Ramesh, A., Sprechmann, P., LeCun, Y.:
  Disentangling factors of variation in deep representation using adversarial
  training. In: Lee, D., Sugiyama, M., Luxburg, U., Guyon, I., Garnett, R.
  (eds.) Advances in Neural Information Processing Systems. vol.~29. Curran
  Associates, Inc. (2016),
  \url{https://proceedings.neurips.cc/paper/2016/file/ef0917ea498b1665ad6c701057155abe-Paper.pdf}

\bibitem{dsprites17}
Matthey, L., Higgins, I., Hassabis, D., Lerchner, A.: dsprites: Disentanglement
  testing sprites dataset. https://github.com/deepmind/dsprites-dataset/ (2017)

\bibitem{miller1995wordnet}
Miller, G.A.: Wordnet: a lexical database for english. Communications of the
  ACM  \textbf{38}(11),  39--41 (1995)

\bibitem{nguyen2021unsupervised}
Nguyen-Meidine, L.T., Belal, A., Kiran, M., Dolz, J., Blais-Morin, L.A.,
  Granger, E.: Unsupervised multi-target domain adaptation through knowledge
  distillation. In: Proceedings of the IEEE/CVF Winter Conference on
  Applications of Computer Vision. pp. 1339--1347 (2021)

\bibitem{niemeyer2021giraffe}
Niemeyer, M., Geiger, A.: Giraffe: Representing scenes as compositional
  generative neural feature fields. In: Proceedings of the IEEE/CVF Conference
  on Computer Vision and Pattern Recognition. pp. 11453--11464 (2021)

\bibitem{oord2018representation}
Oord, A.v.d., Li, Y., Vinyals, O.: Representation learning with contrastive
  predictive coding. arXiv preprint arXiv:1807.03748  (2018)

\bibitem{papernot2016distillation}
Papernot, N., McDaniel, P., Wu, X., Jha, S., Swami, A.: Distillation as a
  defense to adversarial perturbations against deep neural networks. In: 2016
  IEEE symposium on security and privacy (SP). pp. 582--597. IEEE (2016)

\bibitem{pkt_eccv}
Passalis, N., Tefas, A.: Learning deep representations with probabilistic
  knowledge transfer. In: Proceedings of the European Conference on Computer
  Vision (ECCV) (2018)

\bibitem{peters2017elements}
Peters, J., Janzing, D., Sch{\"o}lkopf, B.: Elements of causal inference:
  foundations and learning algorithms. The MIT Press (2017)

\bibitem{quattoni2009recognizing}
Quattoni, A., Torralba, A.: Recognizing indoor scenes. In: 2009 IEEE Conference
  on Computer Vision and Pattern Recognition. pp. 413--420. IEEE (2009)

\bibitem{Sucheng2022CVPR}
Ren, S., Zhou, D., He, S., Feng, J., Wang, X.: Shunted self-attention via
  multi-scale token aggregation. In: Proceedings of the IEEE/CVF Conference on
  Computer Vision and Pattern Recognition (2022)

\bibitem{romero2014fitnets}
Romero, A., Ballas, N., Kahou, S.E., Chassang, A., Gatta, C., Bengio, Y.:
  Fitnets: Hints for thin deep nets. arXiv preprint arXiv:1412.6550  (2014)

\bibitem{ILSVRC15}
Russakovsky, O., Deng, J., Su, H., Krause, J., Satheesh, S., Ma, S., Huang, Z.,
  Karpathy, A., Khosla, A., Bernstein, M., Berg, A.C., Fei-Fei, L.: {ImageNet
  Large Scale Visual Recognition Challenge}. International Journal of Computer
  Vision (IJCV)  \textbf{115}(3),  211--252 (2015).
  \doi{10.1007/s11263-015-0816-y}

\bibitem{sandler2018mobilenetv2}
Sandler, M., Howard, A., Zhu, M., Zhmoginov, A., Chen, L.C.: Mobilenetv2:
  Inverted residuals and linear bottlenecks. In: Proceedings of the IEEE
  conference on computer vision and pattern recognition. pp. 4510--4520 (2018)

\bibitem{silberman2012indoor}
Silberman, N., Hoiem, D., Kohli, P., Fergus, R.: Indoor segmentation and
  support inference from rgbd images. In: European conference on computer
  vision. pp. 746--760. Springer (2012)

\bibitem{Sun2019PatientKD}
Sun, S., Cheng, Y., Gan, Z., Liu, J.: Patient knowledge distillation for bert
  model compression. In: EMNLP (2019)

\bibitem{tian2020makes}
Tian, Y., Sun, C., Poole, B., Krishnan, D., Schmid, C., Isola, P.: What makes
  for good views for contrastive learning? arXiv preprint arXiv:2005.10243
  (2020)

\bibitem{tishby2000information}
Tishby, N., Pereira, F.C., Bialek, W.: The information bottleneck method. arXiv
  preprint physics/0004057  (2000)

\bibitem{tran2017disentangled}
Tran, L., Yin, X., Liu, X.: Disentangled representation learning gan for
  pose-invariant face recognition. In: Proceedings of the IEEE conference on
  computer vision and pattern recognition. pp. 1415--1424 (2017)

\bibitem{tschannen2019mutual}
Tschannen, M., Djolonga, J., Rubenstein, P.K., Gelly, S., Lucic, M.: On mutual
  information maximization for representation learning. arXiv preprint
  arXiv:1907.13625  (2019)

\bibitem{WahCUB_200_2011}
Wah, C., Branson, S., Welinder, P., Perona, P., Belongie, S.: {The Caltech-UCSD
  Birds-200-2011 Dataset}. Tech. Rep. CNS-TR-2011-001, California Institute of
  Technology (2011)

\bibitem{wang2020deep}
Wang, J., Sun, K., Cheng, T., Jiang, B., Deng, C., Zhao, Y., Liu, D., Mu, Y.,
  Tan, M., Wang, X., et~al.: Deep high-resolution representation learning for
  visual recognition. IEEE transactions on pattern analysis and machine
  intelligence  (2020)

\bibitem{yang2020NeurIPS}
Yang, Y., Feng, Z., Song, M., Wang, X.: Factorizable graph convolutional
  networks. In: Conference on Neural Information Processing Systems (2020)

\bibitem{yang2020CVPR}
Yang, Y., Qiu, J., Song, M., Tao, D., Wang, X.: Distilling knowledge from graph
  convolutional networks. In: Proceedings of the IEEE/CVF Conference on
  Computer Vision and Pattern Recognition (2020)

\bibitem{ye2019student}
Ye, J., Ji, Y., Wang, X., Ou, K., Tao, D., Song, M.: Student becoming the
  master: Knowledge amalgamation for joint scene parsing, depth estimation, and
  more. In: Proceedings of the IEEE/CVF Conference on Computer Vision and
  Pattern Recognition. pp. 2829--2838 (2019)

\bibitem{zagoruyko2016wide}
Zagoruyko, S., Komodakis, N.: Wide residual networks. arXiv preprint
  arXiv:1605.07146  (2016)

\bibitem{Zagoruyko2017AT}
Zagoruyko, S., Komodakis, N.: Paying more attention to attention: Improving the
  performance of convolutional neural networks via attention transfer. In: ICLR
  (2017), \url{https://arxiv.org/abs/1612.03928}

\bibitem{zenke2017continual}
Zenke, F., Poole, B., Ganguli, S.: Continual learning through synaptic
  intelligence. In: International Conference on Machine Learning. pp.
  3987--3995. PMLR (2017)

\bibitem{zhang2020side}
Zhang, J.O., Sax, A., Zamir, A., Guibas, L., Malik, J.: Side-tuning: a baseline
  for network adaptation via additive side networks. In: European Conference on
  Computer Vision. pp. 698--714. Springer (2020)

\end{thebibliography}

\end{document}


\pagestyle{headings}
\mainmatter
\def\ECCVSubNumber{5076}  

\title{Factorizing Knowledge in  Neural Networks\\-\textit{Supplementary Materials}-} 


\titlerunning{Abbreviated paper title}
%
\author{Xingyi Yang\orcidlink{https://orcid.org/
0000-0002-1603-9829} \and Jingwen Ye\orcidlink{https://orcid.org/0000-0001-8415-3597} \and Xinchao Wang\orcidlink{https://orcid.org/
0000-0003-0057-1404}}
%
%
\institute{National University of Singapore\\
\email{xyang@u.nus.edu,\{jingweny,xinchao\}@nus.edu.sg}}
\maketitle
In this supplementary, we first provide 
the proof of our variational bound for the
InfoMax-BottleNeck objective, 
and then showcase additional results 
to verify our proposed knowledge factorization. 
Next, we describe our implementation details, 
dataset settings, evaluation metrics, and hyper-parameter settings.
\section{Variational Lower bound for IMB}
\subsection{Lower and Upper bound for Mutual Information
}
For two random variables $X$ and $Y$, mutual information (MI) describes
the independence between each other.
It can be formulated as the Kullback Leibler divergence between the joint probability $p(\mathbf{x}, \mathbf{y})$ 
and the product of marginal distribution $p(\mathbf{x})p(\mathbf{y})$: 

{\footnotesize
\begin{align}
    \mathcal{I}(X,Y) &= D_{KL}[p(\mathbf{x} ,\mathbf{y})||p(\mathbf{x})p(\mathbf{y})] \\
     &=\int \,d \mathbf{x}\,d \mathbf{y} p(\mathbf{x},\mathbf{y})\log \frac{p(\mathbf{y},\mathbf{x})}{p(\mathbf{y})p(\mathbf{x})} \\
     &=\int \,d \mathbf{x} \,d \mathbf{y} p(\mathbf{x},\mathbf{y})\log \frac{p(\mathbf{y}|\mathbf{x})}{p(\mathbf{y})}. \label{eq: mi(z,y)}
\end{align}}\noindent

\textbf{Lower bound.} Since  computing $p(\mathbf{y}|\mathbf{x})$ directly is intractable, we use a variational distribution $q(\mathbf{y}|\mathbf{x})$ to approximate $p(\mathbf{y}|\mathbf{x})$. Because the Kullback Leibler divergence is always positive $D_{KL}[p(\mathbf{y}|\mathbf{x})||q(\mathbf{y}|\mathbf{x})] \geq 0$, we have a lower bound for the MI as: 
{\footnotesize
\begin{align}
    \mathcal{I}(X,Y) &=\int\! \,d \mathbf{x} \,d \mathbf{y} p(\mathbf{x},\mathbf{y})\log \frac{p(\mathbf{y}|\mathbf{x})}{p(\mathbf{y})} \\
    &=\int\! \,d \mathbf{x} \,d \mathbf{y} p(\mathbf{x},\mathbf{y})\log \frac{p(\mathbf{y}|\mathbf{x})q(\mathbf{y}|\mathbf{x})}{p(\mathbf{y})q(\mathbf{y}|\mathbf{x})} \\
    &=\int\! \,d \mathbf{x} \,d \mathbf{y} p(\mathbf{x},\mathbf{y})\log \frac{q(\mathbf{y}|\mathbf{x})}{p(\mathbf{y})}\! +\! D_{KL}[p(\mathbf{y}|\mathbf{x})||q(\mathbf{y}|\mathbf{x})]\\
    &\geq \int \,d \mathbf{x} \,d \mathbf{y} p(\mathbf{x},\mathbf{y})\log \frac{q(\mathbf{y}|\mathbf{x})}{p(\mathbf{y})}\\
    &= \int\! \,d \mathbf{x}\,d \mathbf{y} p(\mathbf{x},\mathbf{y})\log q(\mathbf{y}|\mathbf{x})-  \int\! d \mathbf{y} p(\mathbf{y})\log p(\mathbf{y})  \\
    & = \mathbb{E}_{p(\mathbf{y}, \mathbf{x})}\Big[\log q(\mathbf{y}|\mathbf{x})\Big]+  H(Y) \label{eq8}\\
    & \geq \mathbb{E}_{p(\mathbf{y}, \mathbf{x})}\Big[\log q(\mathbf{y}|\mathbf{x})\Big].
\end{align}
}\noindent
Note that the entropy term $H(Y)\geq 0$ and is sometimes discarded when $H(Y)$ is a constant.

In this work, we assume $q(\mathbf{y}| \mathbf{x})$ to be an energy-based function that is parameterized by a critic function $f(\mathbf{x}, \mathbf{y})$:
{\small\begin{align}
        q(\mathbf{y}|\mathbf{x}) = \frac{p(\mathbf{y})}{C} e^{f(\mathbf{x},\mathbf{y})}, \text{where } C= \mathbb{E}_{p(\mathbf{y})}\big[e^{f(\mathbf{x},\mathbf{y})}\big].
\end{align}}

{We substitute $q(\mathbf{y}|\mathbf{x})$ into Equation~\ref{eq8} and derive an unnormalized lower bound on MI, which we refer to as $I_{\text{UBA}}$ for the Barber and Agakov bound~\cite{agakov2004algorithm} and $I_{\text{DA}}$ Donsker \& Varadhan bound~\cite{donsker1975asymptotic}}. We write:
{\small\begin{align}
        &\mathbb{E}_{p(\mathbf{x}, \mathbf{y})}[\log q(\mathbf{y}| \mathbf{x})] + H(Y) \\\geq&
        \mathbb{E}_{p(\mathbf{x}, \mathbf{y})}[f(\mathbf{x},\mathbf{y})] -  \mathbb{E}_{p( \mathbf{x})}\big[\log C \big]=  I_{\text{UBA}}\\
        \geq& \mathbb{E}_{p(\mathbf{x}, \mathbf{y})}[f(\mathbf{x},\mathbf{y})] -  \log\mathbb{E}_{p( \mathbf{x})}\big[ C\big]=  I_{\text{DV}} \quad [\text{Jensen’s inequality}].
\end{align}}
    
\textbf{Upper bound.} We then consider the upper bound of mutual information. Similarly, we use $q(\mathbf{y})$ as the variational approximation to the marginal distribution of $p(\mathbf{y})$. Using the non-negativity property of KL divergence again, we know that $D_{KL}[p(\mathbf{y})|q(\mathbf{y})] = \int  \,d \mathbf{y} p(\mathbf{y})\log \frac{p(\mathbf{y})}{q(\mathbf{y})}>0$. Therefore we get a tractable variational upper bound for $ \mathcal{I}(X,Y)$:
{\footnotesize
\begin{align}
     \mathcal{I}(X, Y) &=  \int\! \,d \mathbf{x}\,d \mathbf{y} p(\mathbf{x},\mathbf{y})\log \frac{p(\mathbf{y}|\mathbf{x})}{p(\mathbf{y})} \\
     & = \int\! \,d \mathbf{x}\,d \mathbf{y} p(\mathbf{x},\mathbf{y})\log \frac{p(\mathbf{y}|\mathbf{x})q(\mathbf{y})}{p(\mathbf{y})q(\mathbf{y})}\\
     & = \int\! \,d \mathbf{x}\,d \mathbf{y} p(\mathbf{x},\mathbf{y})\log \frac{p(\mathbf{y}|\mathbf{x})}{q(\mathbf{y})}\! -\! \int\!  \,d \mathbf{y} p(\mathbf{y})\log \frac{p(\mathbf{y})}{q(\mathbf{y})}\\
     & \leq \int \,d \mathbf{x}\,d \mathbf{y} p(\mathbf{x},\mathbf{y})\log \frac{p(\mathbf{y}|\mathbf{x})}{q(\mathbf{y})}\\
      & = \int \,d \mathbf{x}\,d \mathbf{y} p(\mathbf{y}|\mathbf{x}) p(\mathbf{x})\log \frac{p(\mathbf{y}|\mathbf{x})}{q(\mathbf{y})}\\
      & = \int \,d \mathbf{x}p(\mathbf{x}) \int\,d \mathbf{y} p(\mathbf{y}|\mathbf{x}) \log \frac{p(\mathbf{y}|\mathbf{x})}{q(\mathbf{y})}\\
     & = \mathbb{E}_{p(\mathbf{x})} \Big[D_{KL}[p(\mathbf{y}|\mathbf{x})||q(\mathbf{y})]\Big].
\end{align}
}
\subsection{Approximate IMB with Variational bounds}
For each task, we assume the three random variables $X, Y_j, T_j$ follow a Markov chain that $X  \rightarrow T_j \rightarrow Y_j$. Because  computing IMB objective directly is intractable, we resort to maximizing a variational lower bound: 
\begin{align*}
    \mathcal{\hat{L}}_{I} &= \mathcal{I}(T_j, Y_j) + \alpha \mathcal{I}(X, Z) - \beta \mathcal{I}(X, T_j) \\
    & \geq \mathbb{E}_{p(\mathbf{y}_j, \mathbf{t}_j)}[\log q(\mathbf{y}_j| \mathbf{t}_j)] + H(Y) \\ & \quad + \alpha \big(\mathbb{E}_{p(\mathbf{z}, \mathbf{x})}[\log q(\mathbf{z}| \mathbf{x})] + H(Z)\big) \\&\quad - \beta \mathbb{E}_{p(\mathbf{x})}\Big[D_{KL}[p(\mathbf{t}_j|\mathbf{x})||q(\mathbf{t}_j)]\Big]\\
    & \geq \mathbb{E}_{p(\mathbf{y}_j, \mathbf{t}_j)}[\log q(\mathbf{y}_j| \mathbf{t}_j)] \\&\quad + \alpha \big(\mathbb{E}_{p(\mathbf{z}, \mathbf{x})}[\log q(\mathbf{z}| \mathbf{x})] + H(Z)\big) \\&\quad- \beta \mathbb{E}_{p(\mathbf{x})}\Big[D_{KL}[p(\mathbf{t}_j|\mathbf{x})||q(\mathbf{t}_j)]\Big],
\end{align*}
where $D_{KL}$ denotes the KL divergence between two distributions and $q(\cdot)$ denotes the variational distributions. The entropy term $H(Y)\geq 0$ is canceled because it is a constant that is irrelevant to the optimization.

\section{Additional Experiments}
In this section, we provide additional experiments to further verify the utility of our proposed KF.
\begin{figure*}[b]

    \centering
    \begin{subfigure}{0.24\linewidth}
        \includegraphics[width=\linewidth]{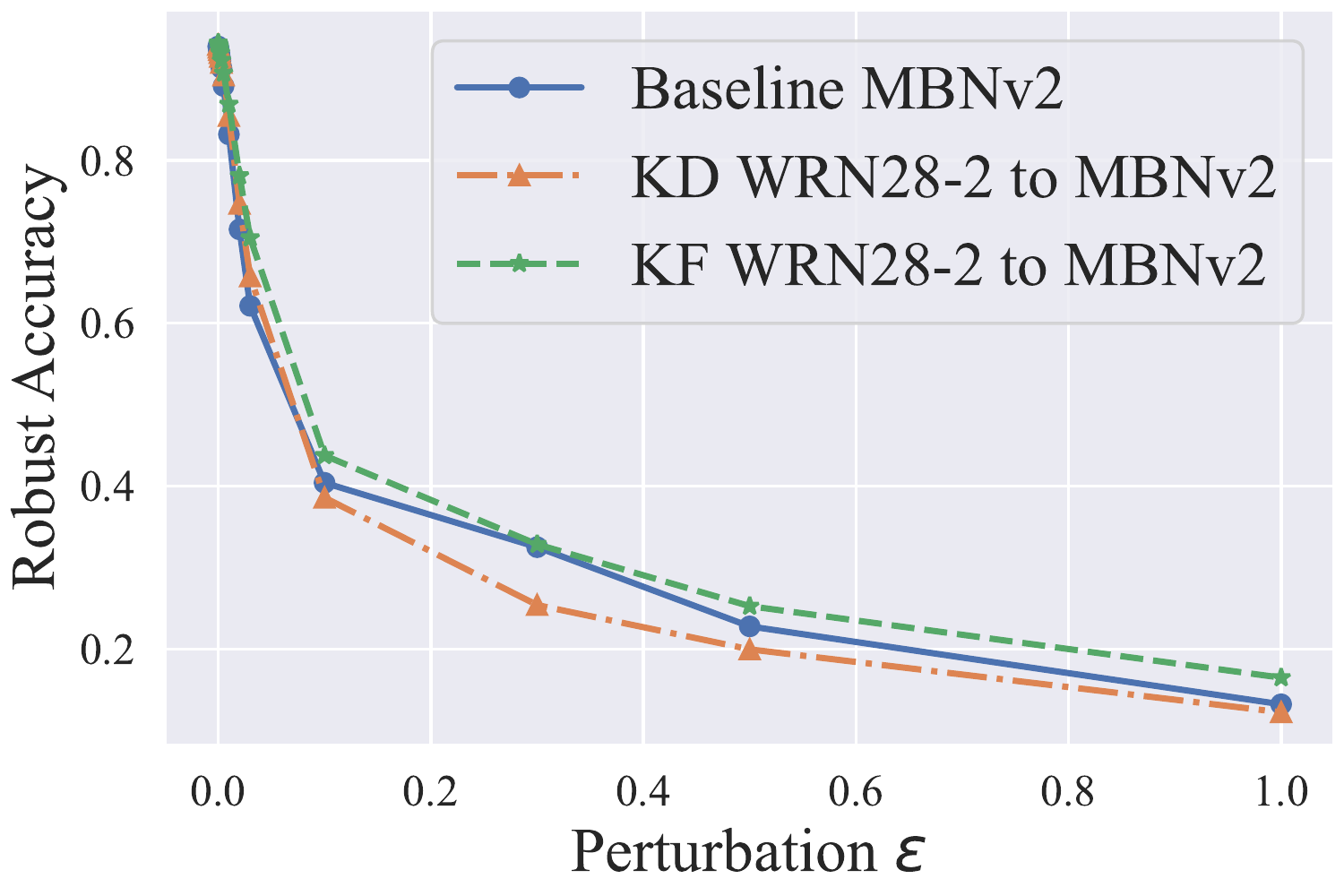}
        \caption{MBNv2 FGSM}
    \end{subfigure}
     \begin{subfigure}{0.24\linewidth}
        \includegraphics[width=\linewidth]{Fig/mbn_PGD(40).pdf}
        \caption{MBNv2 PGD-40}
    \end{subfigure}
    \begin{subfigure}{0.24\linewidth}
        \includegraphics[width=\linewidth]{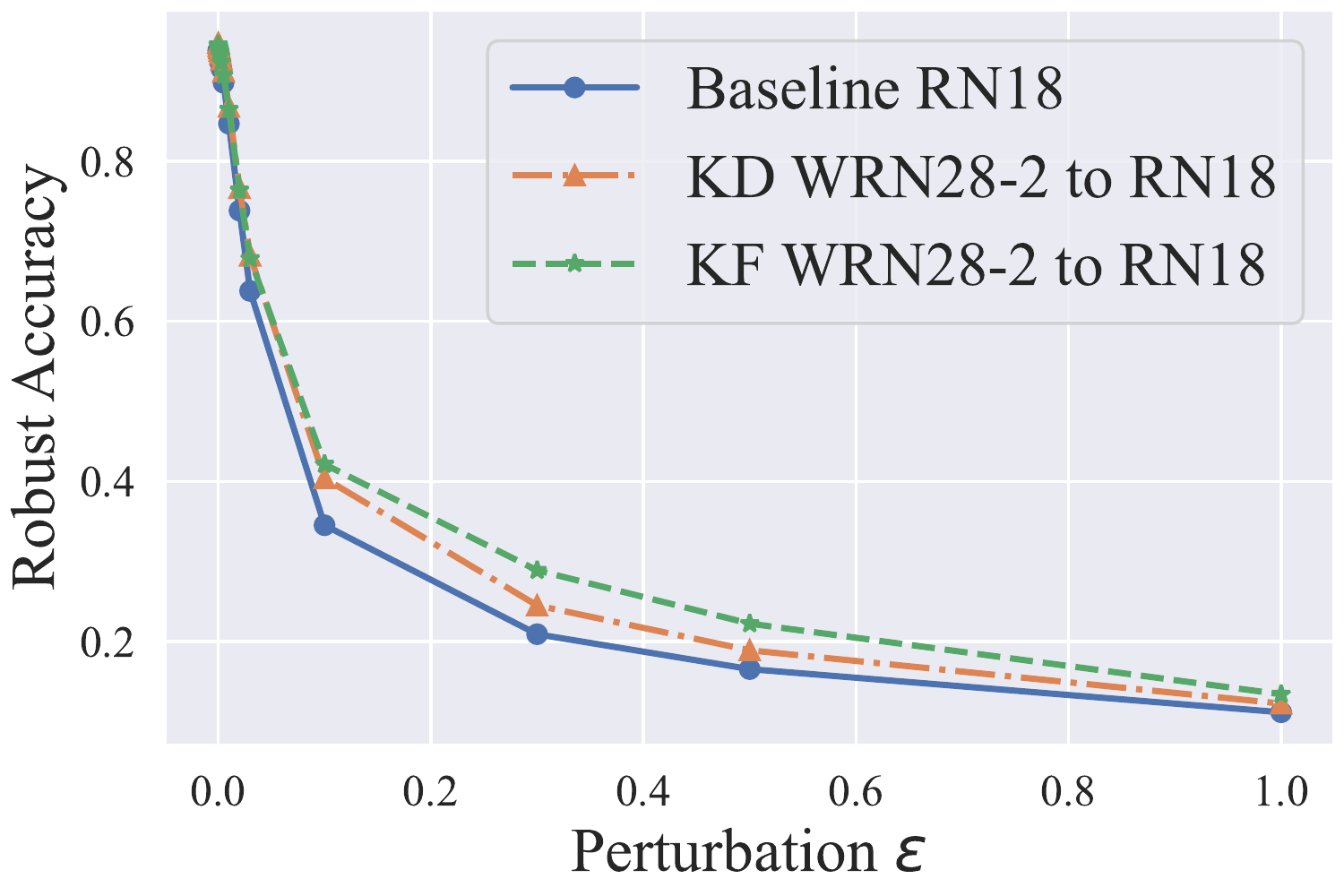}
        \caption{R18 FGSM}
    \end{subfigure}
    \begin{subfigure}{0.24\linewidth}
        \includegraphics[width=\linewidth]{Fig/rn18_PGD(40).pdf}
        \caption{R18 PGD-40}
    \end{subfigure}
    \caption{Adversarial robustness accuracy across backbones and attack. R18 stands for ResNet-18.}
    \label{fig:Adv}
\end{figure*}

\subsection{Factor Networks are Interpretable and Robust}
\begin{figure}
    \centering
    \includegraphics[width=\linewidth]{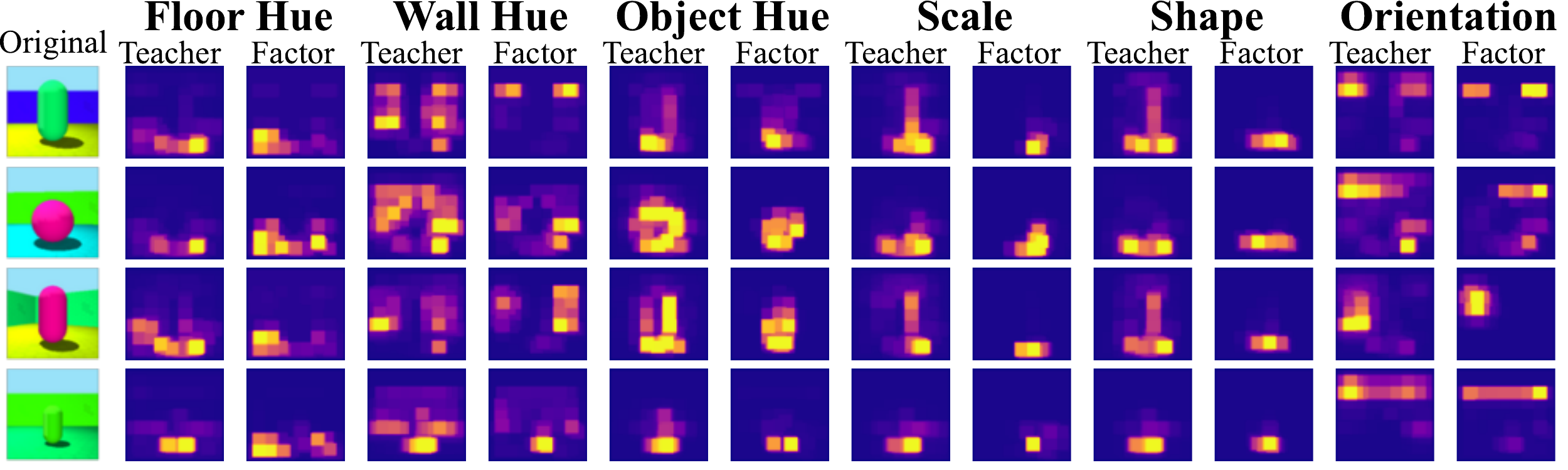}
    \caption{Grad-Cam visualization for ResNet-18 on Shape3D datasets. Given the input in (column 1), we compare the attribution obtained from the teacher network and its factorized students on 6 tasks. }
    \label{fig:Shape3d gradcam}
\end{figure}

\begin{figure}
    \centering
    \includegraphics[width=\linewidth]{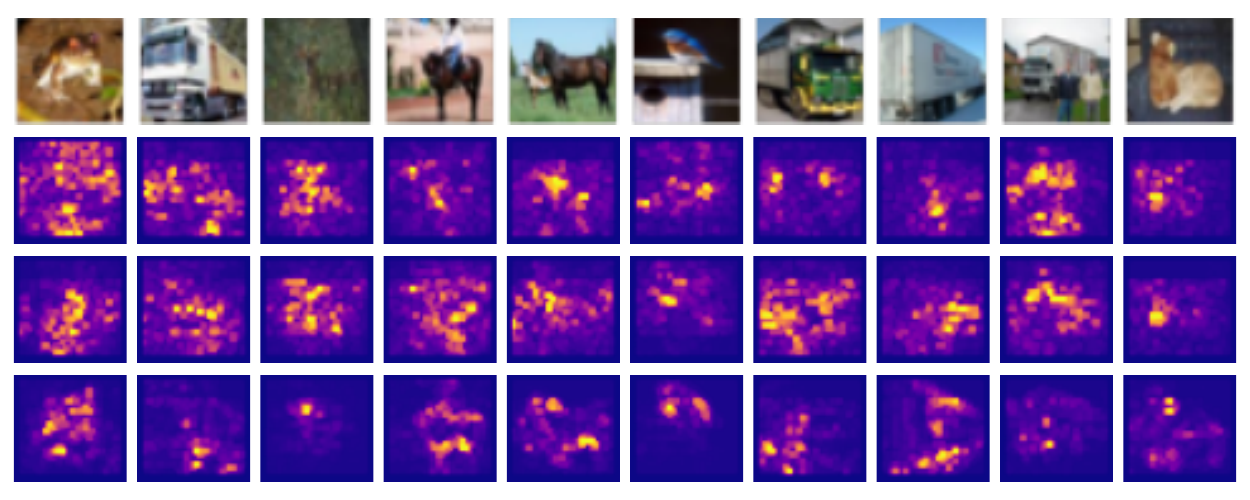}
    \caption{Grad-Cam visualization for MBNv2 on CIFAR-10. (row 1) Input (row 2) Baseline MBNv2 (row 3) MBNv2 distilled from ResNet-18 and (row 4) MBNv2 factorized from ResNet-18.}
    \label{fig:cifar10 gradcam}
\end{figure}

We testify 
if our factorized model 
embraces better interpretability and robustness 
by applying adversarial attacks and 
visualizing its attribution map.\\
\textbf{Robustness Setup.} We apply adversarial perturbations 
to MBNv2 and ResNet-18 to examine their robustness. For each backbone, 
we use three models that are trained with different strategies: (1) Baseline without teacher (2) Distilled from WRN28-2 model and (3) Single-task network factorized from WRN28-2. We apply Fast Gradient Sign Method~(FGSM)~\cite{43405} and Projected Gradient Descent~(PGD)~\cite{madry2018towards} with 40 iterations as our attacks. 
Attacks are conducted with different magnitude $\epsilon$ in $L_{\infty}$ norm ball.\\
\textbf{Attribution Setup.} We apply  Grad-Cam~\cite{selvaraju2017grad} 
to visualize the attribution map 
for ResNet-18 trained on Shape3D 
and MBNv2 trained on CIFAR-10 datasets.
The yellow area highlights 
the most informative region in the image space. 
For each factor network, 
we visualize the Grad-Cam with respect to the last layer of its TSN.\\
\textbf{Robustness and Attribution Results.} Figure~\ref{fig:Adv} illustrates 
the accuracy under different attacks. 
For both attacks and two backbones, 
factor networks~(green dashed plot with star) perform the marginal better than the KD model and the baselines. 
Meanwhile, factor networks provide sharp and noiseless explanation for the input compared with its teacher and KD counterpart, as shown in Figure~\ref{fig:Shape3d gradcam} and~\ref{fig:cifar10 gradcam}. In fact, minimizing $\mathcal{I}(X, T_j)$ can be regarded as posing regularization on task-specific feature $\mathbf{t}^j$. It enforces  $\mathbf{t}^j$ to be compact and sparse, which naturally gives rise to  interpretability to different tasks and robustness to input noise.


\subsection{Sub-Task Integration}

We show here 
how KF performs 
when constructing 
new sub-task prediction models 
by integrating factor networks.

\noindent\textbf{Experiment Setup.} 
We construct 5 binary-class classification tasks
and 5 trinary-class classification tasks from the CIFAR-10.
We deliberately selected some indistinguishable categories 
into a group to increase the challenge.
The binary classification problems include \textit{cat-dog}, \textit{deer-horse}, \textit{automobile-truck}, \textit{bird-airplane} and  \textit{automobile-ship}. 
Each binary classification task contains 10,000 images for training~(5,000 per class) and 1,000 for testing~(500 per class). 
The trinary-class classification problems include \textit{cat-dog-frog}, \textit{deer-horse-dog}, \textit{automobile-truck-ship}, \textit{bird-airplane-frog} and  \textit{automobile-ship-airplane}. Each trinary-class problem contains 15,000 images for training~(5,000 per class) and 1,500 for testing~(500 per class). We compare the classification accuracy for 4 models 
\begin{itemize}
    \item A ResNet-18 trained on full CIFAR-10, 
    \item A MobileNetv2 trained on sub-task without teacher,
    \item A MobileNetv2 trained on sub-task, distilled from full-CIFAR-10 ResNet-18,
    \item Ten factor networks~(MobileNetv2 as CKN, MobileNetv2x0.5 as TSN) trained on full CIFAR-10, factorizied from full-CIFAR-10 ResNet-18. 
\end{itemize}
For the KF experiments, we first factorize 10 factor networks from the teacher model, with each network corresponding to a single category prediction. Then we evaluate the combined prediction of the 2 or 3 factor networks based on the sub-task requirement. We use the soft-target~\cite{Hinton2015DistillingTK} as our knowledge transfer loss function for both KD and KF, with temperature $T=10$. We set the initial learning rate to 0.1, momentum to 0.9, and weight-decay to 0.0001. The models' weights are optimized with SGD for 200 epochs. The learning rate is reduced by 0.1 at the 100-th and 150-th epoch.

\begin{figure*}
    \centering
    \begin{subfigure}{0.45\linewidth}
    \includegraphics[width=\linewidth]{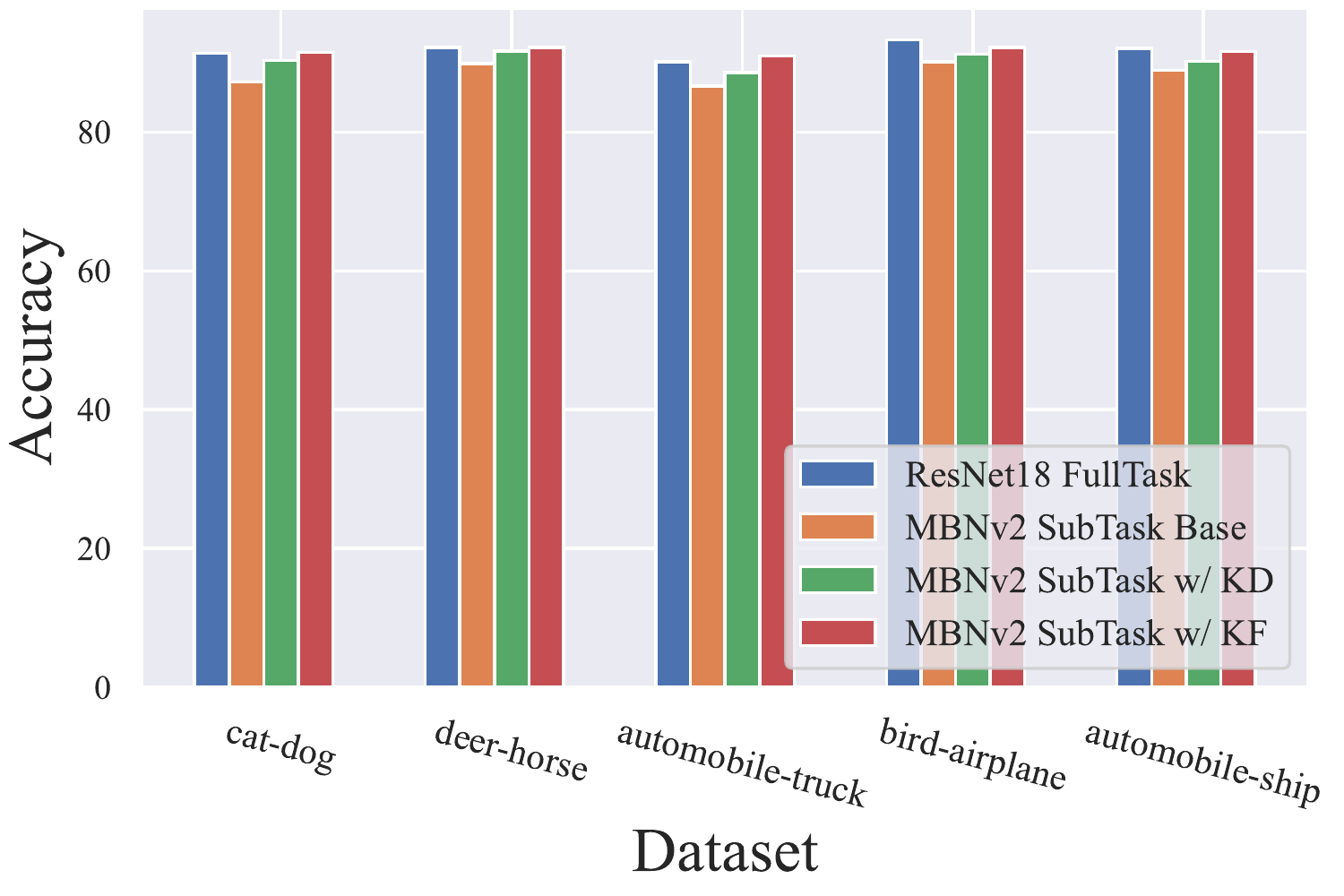}
    \caption{{Binary-class} Classification Task Result}
    \end{subfigure}
    \begin{subfigure}{0.45\linewidth}
    \includegraphics[width=\linewidth]{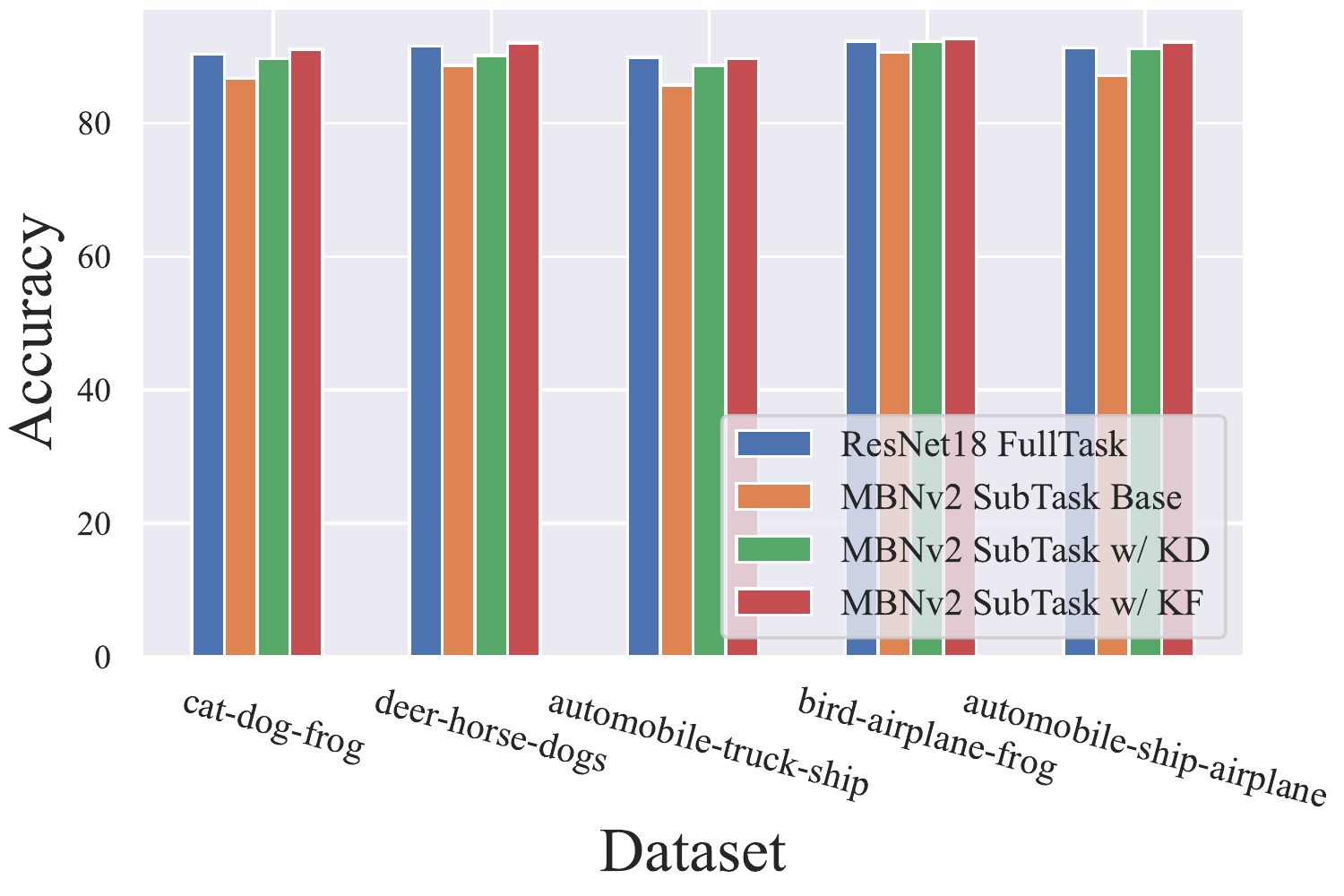}
    \caption{{Trinary-class} Classification Task Result}
    \end{subfigure}
    \caption{Test Accuracy on {binary-class and trinary-class} sub-tasks contrasted on CIFAR-10 with different training strategies.}
    \label{fig:subtask}
\end{figure*}

\noindent\textbf{Results.} Figure~\ref{fig:subtask} provides the test accuracy comparison on 10 CIFAR-10 sub-tasks when trained with different strategies. For all binary-class and trinary-class sub-tasks, we observe that KD considerably improves the baseline MBNv2 accuracy by 2\%. Meanwhile, KF further boosts the performance by 1\%. Ideally, in order to obtain all binary-class and trinary-class sub-task models of CIFAR-10, KD needs to train $\binom{10}{2}=45$ binary-class models and $\binom{10}{3}=120$ trinary-class models, while our proposed KF only needs to train 10 models to be competent for all subtasks. KF significantly reduces training time and model numbers compared with KD.

\subsection{Model Size}
Despite that factorizing a large model into multiple small sub-networks may
{in theory} increase the total number of parameters, we conduct experiments to show that KF, in practise, achieves better performance even with less parameters.

\noindent\textbf{Experiment  Setup.} We compare the number of parameters and the test performance on CIFAR-10. We compare the baseline MBNv2, ResNet-18, WRN-28-2, and WRN-28-10 models and their factor networks with WRN-28-10 in terms of parameter number and accuracy. The experimental setup is the same as the ``Image Classification'' section in the main paper.

\noindent\textbf{Results.} We show the number of parameters and the test performance on CIFAR-10 in Table~\ref{tab:Parameter}. As illustrated, factor network with 1 CKN WRN-28-2 and 1 TSN MBNv2x0.5 improves MBNv2 by 1.22\% with slightly less parameter. Similarly, the factor network with 1 CKN ResNet-18 and 1 TSN MBNv2x0.5 marginally improves WRN-28-10 with only ${2}/{5}$ model parameters. 
These results verify that KF is able to achieve promising test accuracy even with less parameters. 
\begin{table}[]
    \centering
    \small
     \caption{
     Model Parameter Number and Accuracy on CIFAR-10. KF yields better classification accuracy with less parameters.
     }
    \scriptsize
    \label{tab:Parameter}
    \begin{tabular}{l|l|c|c}
    \toprule
        Method & Network & Params(M) & Acc(\%) \\
        \midrule
        Baseline &MBNv2 & 3.50 & 93.58 \\
        KF &CKN WRN-28-2 + 1 TSN MBNv2x0.5 & \textbf{3.44} & \textbf{94.80}\\\midrule
        Baseline &WRN-28-10 & 36.48 & 95.32 \\
        KF &CKN ResNet-18 + 1 TSN MBNv2x0.5 & \textbf{13.66} & \textbf{95.40}\\\midrule
        Baseline &ResNet-18 & 11.69 & 94.45 \\
        KF &CKN WRN-28-2 + 2 TSN MBNv2x0.5 & \textbf{5.41}& \textbf{95.03}\\
        \bottomrule 
    \end{tabular}
\end{table}
\subsection{Visualization for the Multi-task Dense Prediction}
We visualize the prediction results for the multi-task dense prediction on NYUDv2 dataset. We compare 3 methods with ground-truth labels 
\begin{itemize}
    \item Multi-Task HRNet-18, without teacher,
    \item Multi-Task HRNet-18, distilled from Multi-Task HRNet-48,
    \item Multi-Task HRNet-18 as CKN and MBNv2 as TSN, factorized from Multi-Task HRNet-48.
\end{itemize}

\begin{figure*}
    \centering
    \includegraphics[width=\linewidth]{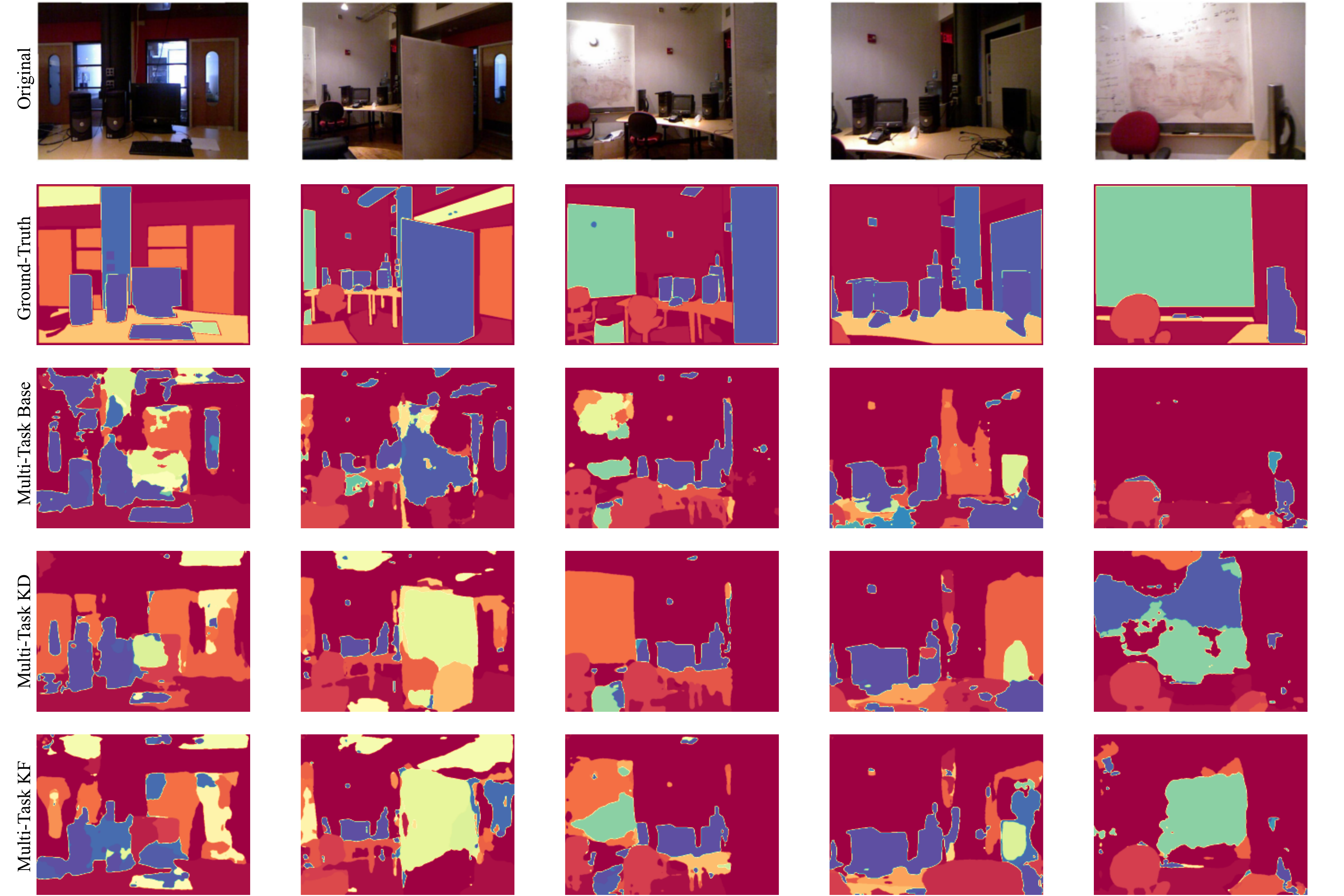}
    \includegraphics[width=\linewidth]{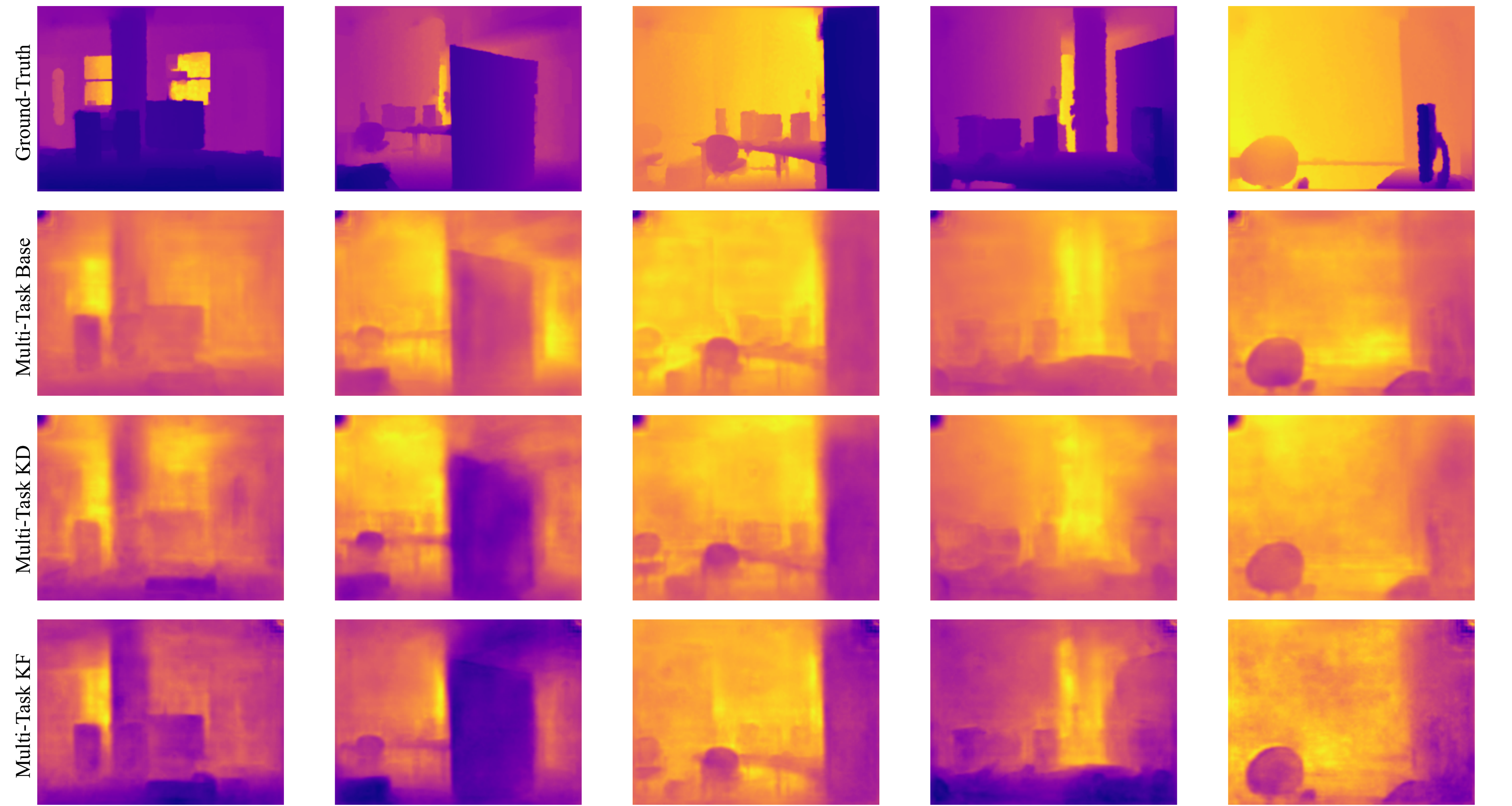}
    \caption{Qualitative results on scene parsing and depth estimation on NYUDv2. }
    \label{fig:nyud vis}
\end{figure*}

\noindent\textbf{Results.} Figure~\ref{fig:nyud vis} shows
the semantic segmentation~(row 2-5) and depth estimation~(row 6-9) 
results on NYUDv2 dataset. We observe that 
factor networks produce more accurate depth estimation 
with sharper edges~(column 2 and 4) compared with the distilled students. On top of this, the knowledge factorization enables students 
to better learn the pattern of the minority categories. For example, the general KD or baseline models 
fail to classify the ``whiteboard'' category, 
which is a rare class in NYUDv2. However, the segmentation 
factor network succeeds in roughly sketching the area of the whiteboard. 

\begin{figure*}
    \centering
    \includegraphics[width=0.95\linewidth]{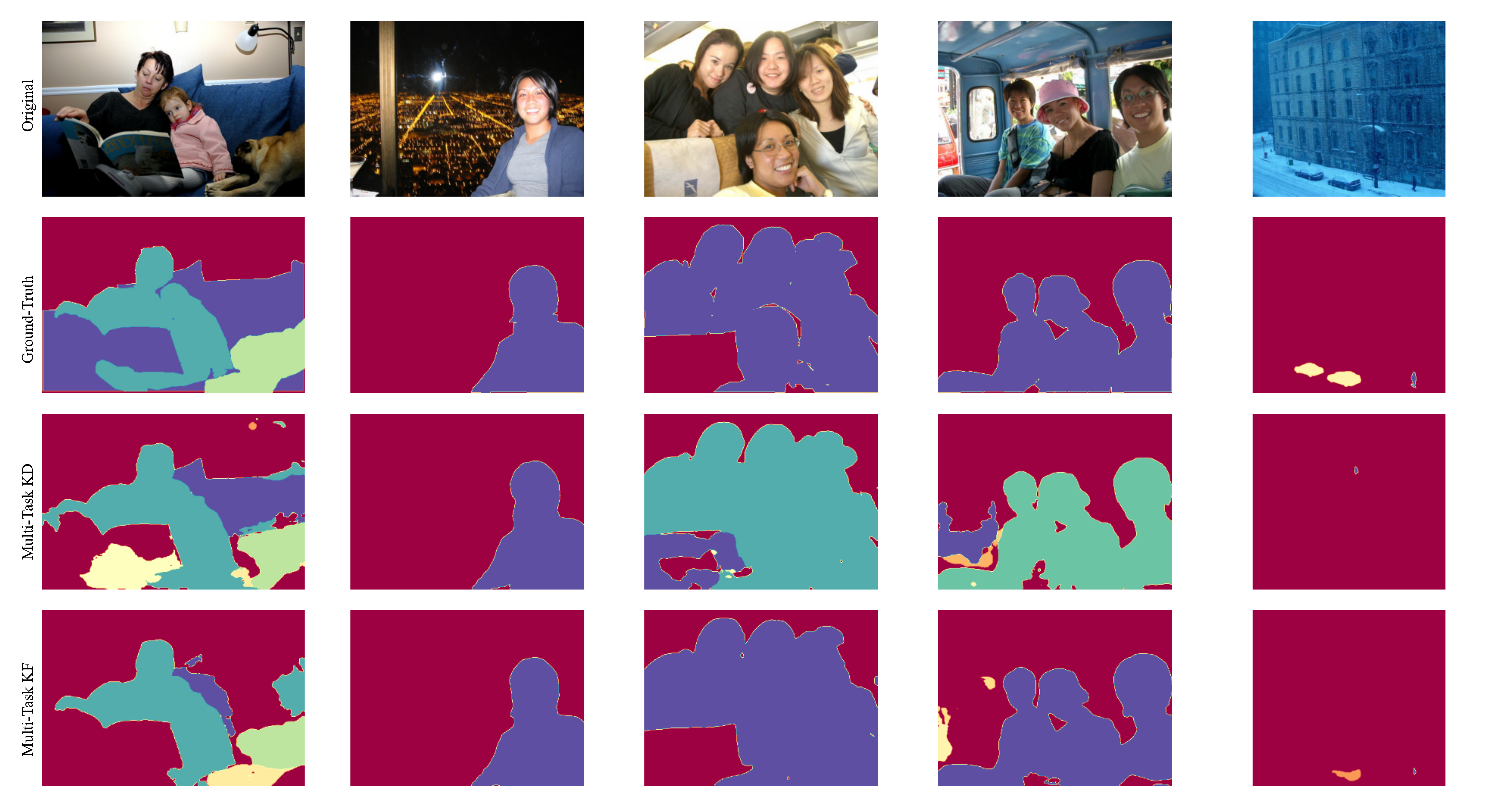}
    \includegraphics[width=0.95\linewidth]{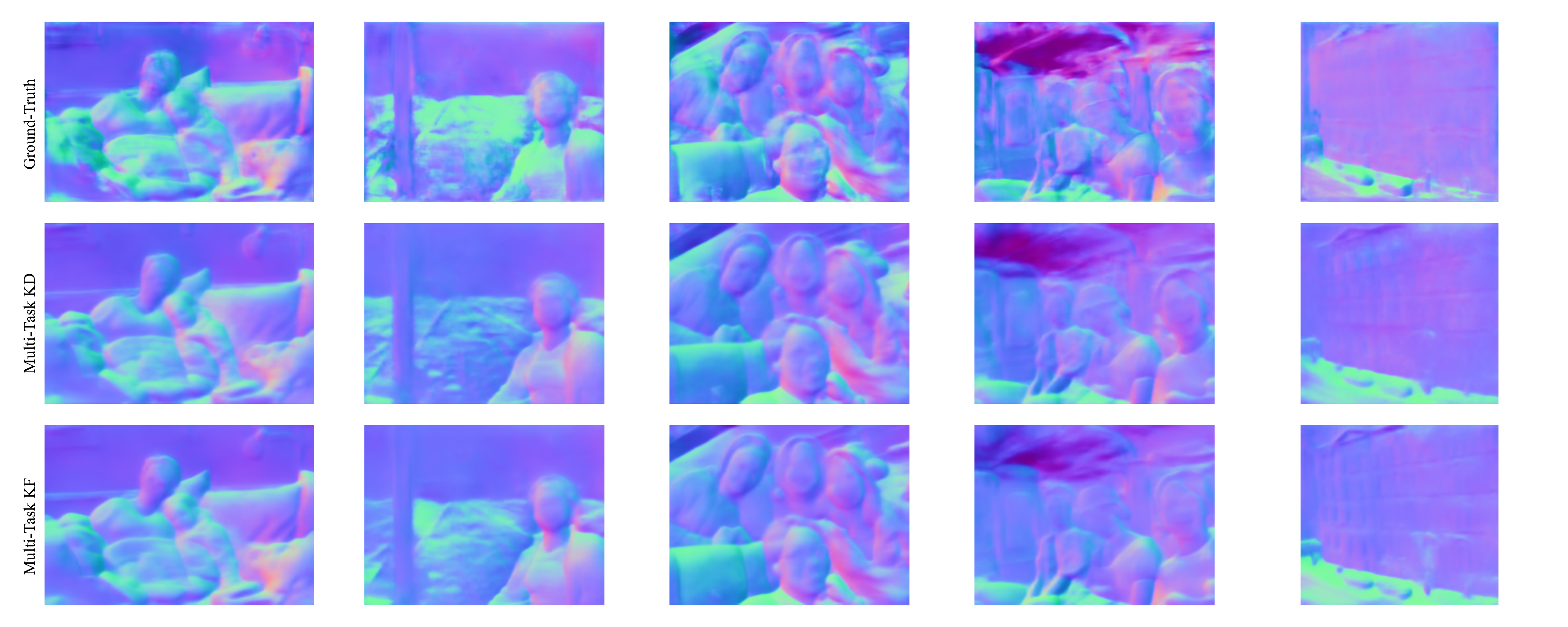}
    \includegraphics[width=0.95\linewidth]{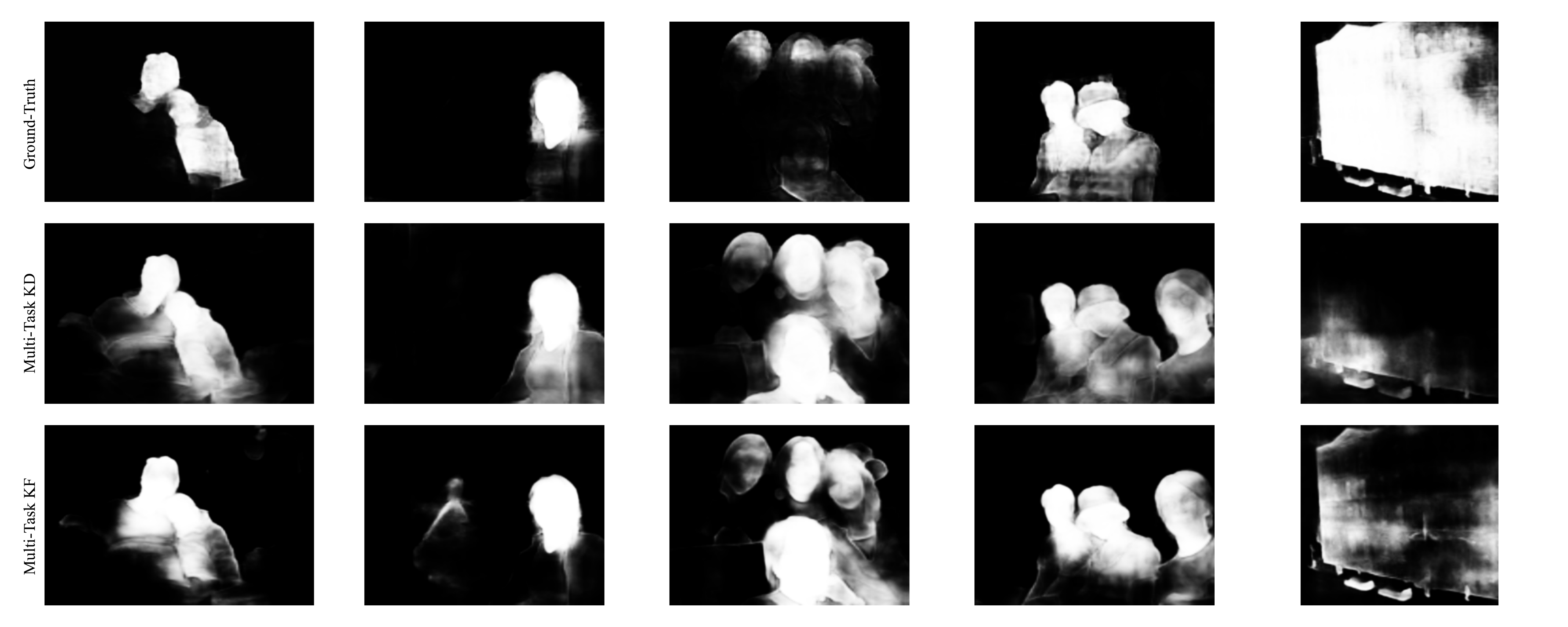}
    \caption{Qualitative results on segmentation, normal estimation, and salient detection on PASCAL-Context dataset. }
    \label{fig:PASCAL vis}
\end{figure*}

Similar results are also visualized for PASCAL-Context in Figure~\ref{fig:PASCAL vis}, in which the factor networks produce more accurate prediction results than those of KD. For example, on the column 2 and 3, the segmentation factor network correctly predicts the ``people'' masks whereas the KD fails. Likewise, the normal-task factor network succeeds to estimate the architectural decoration at the top of the building roof (column 5).

\section{Experiment Details}
\subsection{Datasets} 
\noindent\textbf{ImageNet1K.} ImageNet~\cite{ILSVRC15} is a large-scale image classification dataset. The publicly released dataset contains 1,280,000 training with image-level images with 1,000 object categories. In this work, we also deem it as a 11-task dataset based on the category semantic subtrees. 
The 11 subtrees summarize 11 super-classes: (\texttt{n01466257}, chordate), (\texttt{n01905661}, invertebrate), (\texttt{n02152991},game), (\texttt{n01317541}, domestic animal, domesticated animal), (\texttt{n00021939},artifact, artefact), (\texttt{fa11misc}, Misc), (\texttt{n00019128}, natural object), (\texttt{n09287968}, geological formation, formation), (\texttt{n00007846},person, individual, someone, somebody, mortal, soul), (\texttt{n00017222}, plant, flora, plant life),(\texttt{n12992868}, fungus). 
During training, we apply data augmentation to each sample. We randomly resize and crop a $224\times224$ patch from the image and horizontally flip it with a probability of 0.5. For each test image, we resize it to $256\times 256$ and center-crop a $224\times 224$ patch.

\noindent\textbf{CIFAR10.} The CIFAR-10~\cite{krizhevsky2009learning} dataset consists of 60,000 32x32 colour images in 10 classes, with 6,000 images per class. There are 50,000 training images and 10,000 test images. 
{In this work, we deem it as a 10-task dataset, with 10 binary classification tasks on each category, and also as a 2-task dataset, with a vehicle super-class~(airplane, automobile, ship, truck) and an animal super-class~(bird, cat, deer, dog, frog, horse).}
Each training sample is randomly cropped to a $32\times32$ sample and randomly flipped with a probability of 0.5. We do not apply any data augmentation on the test set.

\noindent\textbf{dSprites.}
dSprites~\cite{dsprites17} is a dataset of 2D shapes procedurally generated from 6  independent latent factors. These factors are \textit{color}, \textit{shape}, \textit{scale}, \textit{rotation}, \textit{x positions} and \textit{y positions} of a sprite. The latent factor values are shown in Table~\ref{tab:Latent dSprites}. All possible combinations of these latents variable are present exactly once, generating N = 737,280 total images. We do not apply any data augmentation when training or testing on dSprites.
\begin{table}[]
    \centering
    \begin{tabular}{c|c}
    \toprule
        Latent factor & Value \\
        \midrule
        Color & White\\
        Shape & square, ellipse, heart\\
        Scale & 6 values linearly spaced in [0.5, 1]\\
        Orientation& 40 values in [0, 2$\pi$]\\
        Position X& 32 values in [0, 1]\\
        Position Y& 32 values in [0, 1]\\
        \bottomrule
    \end{tabular}
    \caption{Latent factor values for dSprites Dataset}
    \label{tab:Latent dSprites}
\end{table}

\noindent\textbf{Shape3D.} Shape3D~\cite{3dshapes18} is a dataset of 3D shapes procedurally generated from 6 independent latent factors. These factors are \textit{floor colour}, \textit{wall colour}, \textit{object colour}, \textit{scale}, \textit{shape}, and \textit{orientation}. The latent factor values are shown in Table~\ref{tab:Latent Shape3d}. All possible combinations of these latents are present exactly once, generating N = 480,000 total images. Each image is of the scale $64\times64$. We do not apply any data augmentation when training or testing on Shape3D.
\begin{table}[]
    \centering
    \begin{tabular}{c|c}
    \toprule
        Latent factor & Value \\
        \midrule
        floor hue & 10 values linearly spaced in [0, 1]\\
        wall hue & 10 values linearly spaced in [0, 1]\\
        object hue & 10 values linearly spaced in [0, 1]\\
        scale & 8 values linearly spaced in [0, 1]\\
        shape & 4 values in [0, 1, 2, 3]\\
        orientation & 15 values linearly spaced in [-30, 30]\\
        \bottomrule
    \end{tabular}
    \caption{Latent factor values for Shape3D Dataset}
    \label{tab:Latent Shape3d}
\end{table}

\noindent\textbf{NYU-Depth V2.} The NYU-Depth V2~(NYUDv2)~\cite{silberman2012indoor} dataset comprises  video sequences from a variety of indoor scenes as recorded by both the RGB and Depth cameras from the Microsoft Kinect. It consist of 795 training and 654 testing images of indoor scenes, with annotations for 40-class semantic segmentation (``Seg.''), depth estimation (``Depth.''), surface normal estimation and boundary detection. We only include the semantic segmentation and depth estimation tasks in our implementation. Each training sample is horizontally flipped with a probability of 0.5 and re-scaled by a factor of $s \in \{1.0, 1.2, 1.5\}$. Finally, the training images are resized to $480 \times 640$. All the test images are also resized to $480 \times 640$, without other data augmentations.

\noindent\textbf{PASCAL-Context.} The PASCAL-Context~\cite{chen2014detect} is a split of the larger PASCAL dataset~\cite{chen2014detect}, providing 4,998 training and 5,105 testing images, labeled for 20-class semantic segmentation (``Seg.''), human parts segmentation~(``H.Part.''), saliency estimation (``Sal.''), surface normal estimation~(``Norm.''), and boundary detection. We only include the first 4 tasks in our work. Each training sample is horizontally flipped with a probability of 0.5, randomly rotated with a degree sampled from $\theta \in [-20^{\circ},20^{\circ}]$ and re-scaled by a factor of $s \in [0.75, 1.25]$. Finally, the training images are resized to $512 \times 512$. All the test images are also resized to $512 \times 512$, without other data augmentations.

\noindent\textbf{CUB-200-2011.} CUB-200-2011~\cite{WahCUB_200_2011} is an extended version of the CUB-200 dataset~\cite{WelinderEtal2010}, with roughly double the number of images per class and new part location annotations. The dataset contains 11,788 images of 200 bird species,  including 5,994 training samples. Each training sample is randomly cropped to $256\times 256$ and randomly horizontal flipped with the probability of 0.5 for augmentation. 

\noindent\textbf{Indoor Scene Dataset.} The MIT-67 Indoor Scene Recognition Dataset is an indoor scene imagary classification dataset. It has 15,620 images in total amongst 67 classes including  airport, train station, kitchen, and library. We randomly select 70\% for training and the rest for testing. Each training sample is randomly flipped with a probability of 0.25 horizontal and 0.25 vertical, randomly rotated with a degree $\theta \in \{90^{\circ},180^{\circ},270^{\circ}\}$, and randomly cropped to $224\times 224$. 

\subsection{Feature Similarity} On dsprites and Shape3D, we utilize a 6-layer CNN as teacher and 3-layer CNNs for the rest models. On CIFAR10, we report results with a ResNet18 teacher and ResNet18 students/CKN, alongside with MBNv2x0.5 TSNs. On NYUDv2, we adopt the ResNet50 teacher, and ResNet18 students or CKA, with MBNv2x0.5 TSNs.
\subsection{Common Knowledge Network and Task specific network Implementation} 
As mentioned in the main paper, each factor network is composed of two network modules, namely Common Knowledge Network~(CKN) and Task Specific Network~(TSN). 

\noindent\textbf{TSN.} Based on the derivation for the IMB bound, $\mathbf{t}_j$ is stochastically sampled from a Gaussian distribution parameterized by the TSN. Specifically, the channel-wise mean and variance of the output feature vector of a TSN is computed. At the same time, $\mathbf{t}_j$ should be powerful to make the individual task prediction with another task head $\mathcal{H}'_{T_j}$:
\begin{align}
     \mathbf{t}_j = \mathcal{S}_{T_j}(\mathbf{x}; \Theta_{\mathcal{S}_{T_j}});\bm\mu_{t_j}=\mathbb{E}[\mathbf{t}_j], \bm \sigma^2_{t_j}=\text{Var}[\mathbf{t}_j],\\
     \hat{y}'_j = \mathcal{H}'_{j}(\mathbf{t}_j; \Theta_{\mathcal{H}'_{j}}).
\end{align}

\noindent\textbf{Final Prediction.} Each input data is fed into two networks in parallel and two feature vectors $\mathbf{z}$ and $\mathbf{t}_j$ from the last layer of CKN and TSN are added as the full task representation. 
The final task prediction is made by the full task representation:
\begin{align}
    \mathbf{z} = \mathcal{S}_C(\mathbf{x}; \Theta_{\mathcal{S}_{C}}); \mathbf{t}_j = \mathcal{S}_{T_j}(\mathbf{x}; \Theta_{\mathcal{S}_{T_j}}),\\
    \hat{y}_j = \mathcal{H}_j(\mathbf{z}+ \mathbf{t}_j; \Theta_{\mathcal{H}_{j}}).
\end{align}

\subsection{Implementation Details and Hyper-parameters}

\noindent\textbf{Disentanglement Experiments.} 
We optimize the model with the following hyper-parameter settings:
\begin{itemize}
    \item We adopt the Adam~\cite{kingma2014adam} method, with initial learning rate set to be 1e-4, weight decay to be 1e-4, and mini-batch size to be 128. 
    \item We train the networks for 20 and 5 epochs on dSprites and 3dshapes respectively. The learning rate is reduced by 0.1 at 10-th and 15-th epoch on dSprites.
    \item The network structure is shown in Table~\ref{tab:disentangle net}.
\end{itemize}   
We use the evaluation code from \textit{disentanglement\_lib}\footnote{https://github.com/google-research/disentanglement\_lib}.

\begin{table}[]
    \centering
    \begin{tabular}{l|c}
    \toprule
        Layer & Parameter \\
        \midrule
        Input & $64\times64\times c$\\
        Conv-ReLU & $4\times4$, $c=32$, stride$=2$ \\
        Conv-ReLU & $4\times4$, $c=32$, stride$=2$ \\
        Conv-ReLU & $4\times4$, $c=64$, stride$=2$ \\
        Conv-ReLU & $4\times4$, $c=128$, stride$=2$ \\
        Conv-ReLU & $4\times4$, $c=256$, stride$=2$ \\
        Conv-ReLU & $4\times4$, $c=256$, stride$=2$ \\
        FC & $c=10$\\
        \bottomrule
    \end{tabular}
    \caption{6-layer network architecture for the disentanglement experiment.}
        \begin{tabular}{l|c}
    \toprule
        Layer & Parameter \\
        \midrule
        Input & $64\times64\times c$\\
        Conv-ReLU & $4\times4$, $c=32$, stride$=2$ \\
        Conv-ReLU & $4\times4$, $c=32$, stride$=2$ \\
        Conv-ReLU & $4\times4$, $c=64$, stride$=2$ \\
        Global Avg-Fool & to $1\times 1$\\
        FC & $c=10$\\
        \bottomrule
    \end{tabular}
    \caption{3-laye network architecture for the disentanglement experiment.}
    \label{tab:disentangle net}
\end{table}

\noindent\textbf{Image Classification Experiments.} 
\begin{itemize}
    \item On CIFAR-10, we adopt SGD to optimize the objectives, 
    with 0.1 initial learning rate and a momentum term of 0.9. 
    We train the network for 200 epochs and 
    the learning rate is reduced by 0.1 at 100-th and 150-th epoch.
    \item On ImageNet1K, we adopt SGD to train the model and set the initial learning rate to 0.1 for ResNet-students and 0.05 for MBNv2, with cosine annealing policy and batch-size of 256. The networks are trained for 150 epochs.
    \item The knowledge transfer loss is set to a soft-target loss with temperature $T=10$. We use binary cross-entropy as our supervised loss for all experiments.
\end{itemize}  The classification implementation is based on the \textit{mmclassification}\footnote{https://github.com/open-mmlab/mmclassification} framework.

\noindent\textbf{Multi-Task Dense Experiments.} 
\begin{itemize}
    \item On ResNet-18 and ResNet-50 backbone, we use the Deeplabv3~\cite{chen2017rethinking} as our decoder. For the HRNet backbone, we concatenate the (upsampled) representations that are from all the resolutions to make the final prediction~\cite{wang2020deep}.
    \item We use the Adam~\cite{kingma2014adam} optimizer to train the model, with initial learning rate of 1e-4, weight decay of 1e-4, and batch-size of 12. The learning rate is updated using polynomial policy. We train all models for 80 epochs on NYUDv2 and 100 epochs on PASCAL-Context.
    \item All network are trained to minimize their cross-entropy loss on semantic segmentation task or L1 distance on other tasks with respect to the ground-truth labels. All task weights are set to be 1.
    \item We distill the knowledge from pretrained teacher network by minimizing the feature difference between the teacher and students with a dense L1 norm function.
\end{itemize}
    The multi-task dense prediction is implemented base on the \textit{Multi-Task-Learning-PyTorch}\footnote{https://github.com/SimonVandenhende/Multi-Task-Learning-PyTorch}
open source repository.

\noindent\textbf{Attribution Experiments.} 
We visualize the Grad-CAM map by using the pretrained classification model on CIFAR-10 and 3dshapes. 

\begin{itemize}
    \item On the 3dshapes experiments with ResNet-18, we visualize the Grad-CAM with respect to the output of layer \textsc{layer4.relu}.
    \item On the CIFAR-10 experiments with MBNv2, we visualize the Grad-CAM that corresponds to the layer \textsc{conv2.bn}, which is the last batch-norm layer of the MBNv2.
\end{itemize} The Grad-CAM is implemented base on the \textit{Captum}\footnote{https://captum.ai} library.

\noindent\textbf{Adversarial Experiments.} We apply 2 adversarial attacks to 
the model trained on CIFAR-10 to testify their robustness to adversarial samples. 
\begin{itemize}
    \item The attack magnitude $\epsilon$ is defined as the relative pixel perturbation level $k/255$.
    \item We transverse all $\epsilon \in \{0.0,
    0.0005,0.001,0.001 5,$
    $ 0.002, 0.003, 0.005, 0.01, 0.02$\\$, 0.03, 0.1, 0.3, 0.5, 1.0\}$ and report the test accuracy.
\end{itemize}
 All attacks are implemented using the open-sourced \textit{foolbox}\footnote{https://github.com/bethgelab/foolbox} library.

\noindent\textbf{Transfer Learning Experiments.} On both CUB-2011-bird and Indoor Scene dataset, we use stochastic gradient descent with momentum of 0.9 and learning rate of 0.1 for 200 epochs for training both from scratch and with distillation. For applying distillation, we set $T= 20$.
\subsection{Disentanglement Metrics} 
In the main paper, we utilize 4 data evaluation metrics to quantify how well the learned representations disentangle with latent factors. We introduce the definition of each metrics in detail.

\noindent\textbf{FactorVAE.} FactorVAE~\cite{Kim2018DisentanglingBF} metric measures disentanglement as the accuracy of a majority vote classifier on a different feature vector that predicts the index of a fixed factor of variation. 

\noindent\textbf{Mutual Information Gap.} For each {ground-truth factor of variation}, the Mutual Information Gap
(MIG)~\cite{chen2018isolating} measures  
the mutual information difference between the top two latent variables with highest mutual information. In our implementation, we instead compute the mutual information difference between two feature dimension with highest MI.

\noindent\textbf{Disenanglement-Completness-Informativeness (DCI).}
DCI~\cite{eastwood2018framework} computes the entropy of the distribution obtained by normalizing the importance of each dimension of the learned representation for predicting the value of a factor of variation. We only include the disentanglement score in our evaluations.

\noindent\textbf{SAP.} The SAP score~\cite{kumar2017variational} is the average difference of the prediction error of the two most predictive latent dimensions for each factor.

\bibliographystyle{splncs04}
\bibliography{egbib}